\documentclass[10pt,twocolumn,letterpaper]{article}

\usepackage{cvpr}
\usepackage{times}
\usepackage{epsfig}
\usepackage{graphicx}
\usepackage{amsmath}
\usepackage{amssymb}
\usepackage{subfigure}
\usepackage{enumerate}
\usepackage{epstopdf}

\usepackage{amsmath}
\usepackage{amssymb}
\usepackage{multirow}

\usepackage[]{algorithm}
\usepackage{algorithmic}

\def\Fig#1{{Fig.\ \ref{fig:#1}}}
\def\Eq#1{{Eq.\ \ref{eq:#1}}}
\def\Tbl#1{{Table \ref{tbl:#1}}}
\def\Sec#1{{Sec.\ \ref{sec:#1}}}

\usepackage[pagebackref=true,breaklinks=true,letterpaper=true,colorlinks,bookmarks=false]{hyperref}

\cvprfinalcopy 

\def\cvprPaperID{2986} 

\ifcvprfinal\fi

\begin{document}

\renewcommand{\dblfloatpagefraction}{1}
\renewcommand{\textfraction}{0}
\renewcommand{\dblfloatsep}{5pt}
\renewcommand{\dbltextfloatsep}{1ex}
\renewcommand{\dbltopfraction}{1}
\renewcommand{\intextsep}{5pt}
\renewcommand{\floatpagefraction}{1}
\renewcommand{\floatsep}{5pt}
\renewcommand{\textfloatsep}{1ex}
\renewcommand{\topfraction}{1}
\renewcommand{\abovecaptionskip}{2pt}
\let\OldCaption=\caption
\renewcommand{\caption}[1]{\small\OldCaption{\em#1}}

\title{Deep Lesion Graphs in the Wild: Relationship Learning and Organization of Significant Radiology Image Findings in a Diverse Large-scale Lesion Database}

\author{Ke Yan, Xiaosong Wang, 
	Le Lu, Ling Zhang, 
	Adam P.\ Harrison \\
	Mohammadhadi Bagheri,
	Ronald M.\ Summers\\
	Imaging Biomarkers and Computer-Aided Diagnosis Laboratory \\National Institutes of Health Clinical Center, 
	10 Center Drive, Bethesda, MD 20892\\
	{\tt\small \{ke.yan, xiaosong.wang, ling.zhang3, mohammad.bagheri, rms\}@nih.gov,}\\ {\tt\small\{lel, aharrison\}@nvidia.com}
}

\maketitle

\begin{abstract}
Radiologists in their daily work routinely find and annotate significant abnormalities on a large number of radiology images. Such abnormalities, or lesions, have collected over years and stored in hospitals' picture archiving and communication systems. However, they are basically unsorted and lack semantic annotations like type and location. In this paper, we aim to organize and explore them by learning a deep feature representation for each lesion. A large-scale and comprehensive dataset, DeepLesion, is introduced for this task. DeepLesion contains bounding boxes and size measurements of over 32K lesions. To model their similarity relationship, we leverage multiple supervision information including types, self-supervised location coordinates, and sizes. They require little manual annotation effort but describe useful attributes of the lesions. Then, a triplet network is utilized to learn lesion embeddings with a sequential sampling strategy to depict their hierarchical similarity structure. Experiments show promising qualitative and quantitative results on lesion retrieval, clustering, and classification. The learned embeddings can be further employed to build a lesion graph for various clinically useful applications. An algorithm for intra-patient lesion matching is proposed and validated with experiments.

\end{abstract}

\section{Introduction}
\label{sec:intro}

Large-scale datasets with diverse images and dense annotations \cite{Feifei09ImageNet, everingham2010pascal, lin2014microsoft} play an important role in computer vision and image understanding, but often come at the cost of vast amounts of labeling. In computer vision, this cost has spurred efforts to exploit weak labels \cite{zhang2016online, krause2016unreasonable, chen2015webly}, \eg{}, the enormous amount of weak labels generated everyday on the web. A similar situation exists in the medical imaging domain, except that annotations are even more time consuming and require extensive clinical training, which precludes approaches like crowd-sourcing. Fortunately, like web data in computer vision, a vast, loosely-labeled, and largely untapped data source does exist in the form of hospital picture archiving and communication systems (PACS). These archives house patient images and accompanying radiological reports, markings, and measurements performed during clinical duties. However, data is typically unsorted, unorganized, and unusable in standard supervised machine learning approaches. Developing means to fully exploit PACS radiology database becomes a major goal within the field of medical imaging. 

This work contributes to this goal of developing an approach to usefully mine, organize, and learn the relationships between lesions found within computed tomography (CT) images in PACS. Lesion detection, characterization, and retrieval is an important task in radiology \cite{Wang2017zoomin, esteva2017dermatologist, Litjens2017survey, Li2018review}. The latest methods based on deep learning and convolutional neural networks (CNNs) have achieved significantly better results than conventional hand-crafted image features \cite{Greenspan2016guest, Litjens2017survey}. However, large amounts of training data with high quality labels are often needed. To address this challenge, we develop a system designed to exploit the routine markings and measurements of significant findings that radiologists frequently perform \cite{Eisen09RECIST}. These archived measurements are potentially highly useful sources of data for computer-aided medical image analysis systems. However, they are basically unsorted and lack semantic labels, \eg{}, lung nodule, mediastinal lymph node. 

We take a feature embedding and similarity graph approach to address this problem. First, we present a new dataset: DeepLesion\footnote{Available at \url{https://nihcc.box.com/v/DeepLesion}.}, which was collected from the PACS of a major medical institute. It contains 32,120 axial CT slices from 10,594 CT imaging studies of 4,427 unique patients. There are 1--3 lesions in each image with accompanying bounding boxes and size measurements. The lesions are diverse but unorganized. Our goal is to understand them and discover their relationships. In other words, can we organize them so that we are able to (1) know their type and location; (2) find similar lesions in different patients, \ie{}, content-based lesion retrieval; and (3) find similar lesions in the same patient, \ie{}, lesion instance matching for disease tracking?

\begin{figure*}[t]
	\begin{center}
		\includegraphics[width=\linewidth,trim=0 0 0 0, clip]{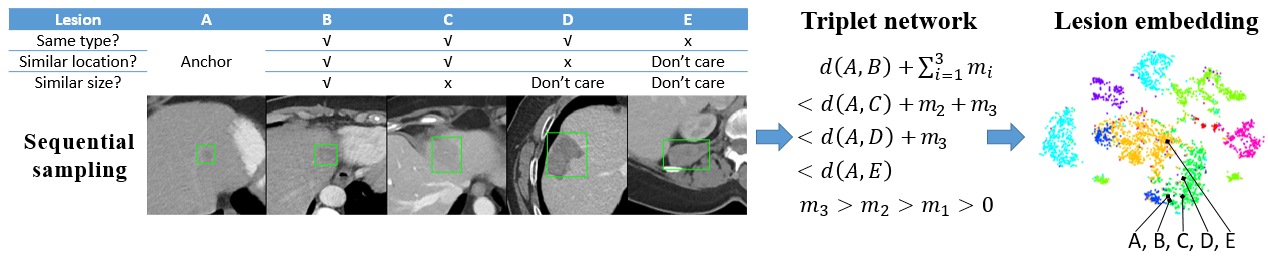} 
	\end{center}
	\caption{The proposed framework. Using a triplet network, we learn a feature embedding for each lesion in our comprehensive DeepLesion dataset. Training samples $ A$--$E $ are selected with a sequential sampling strategy so as to make the embeddings respects similarity in type, location, and size. }
	\label{fig:framework}
\end{figure*}

As \Fig{framework} illustrates, the above problems can be addressed by learning feature representations for each lesion that keeps a proper similarity relationship, \ie{}, lesions with similar attributes should have similar embeddings. To reduce annotation workload and leverage the intrinsic structure within CT volumes, we use three weak cues to describe each lesion: type, location, and size. Lesion types are obtained by propagating the labels of a small amount of seed samples to the entire dataset, producing pseudo-labels. The 3D relative body location is obtained from a self-supervised body-part regression algorithm. Size is directly obtained by the radiological marking. We then define the similarity relationship between lesions based on a hierarchical combination of the cues. A triplet network with a sequential sampling strategy is utilized to learn the embeddings. We also apply a multi-scale multi-crop architecture to exploit both context and detail of the lesions, as well as an iterative refinement strategy to refine the noisy lesion-type pseudo-labels.

Qualitative and quantitative experimental results demonstrate the efficacy of our framework for several highly important applications. {\bf 1)}, we show excellent performance on content-based lesion retrieval \cite{Ramos2016content, Wei2016sim, tsochatzidis2017computer, Li2018review}. Effective solutions to this problem can help identify similar case histories, better understand rare disorders, and ultimately improve patient care \cite{Litjens2017survey}. We show that our embeddings can be used to find lesions similar in type, location, and size. Most importantly, the embeddings can match lesions with semantically similar body structures that are not specified in the training labels. {\bf 2)}, the embeddings are also successfully applied in intra-patient lesion matching. Patients under therapy typically undergo CT examinations (studies) at intervals to assess the effect of treatments. Comparing lesions in follow-up studies with their corresponding ones in previous studies constitutes a major part of a radiologist's workload \cite{moltz2012workflow}. We provide an automated tool for lesion matching which can significantly save time, especially for patients with multiple follow-up studies \cite{sevenster2015improved}.

\section{Related work}
\label{sec:relWork}

{\bf Deep Metric Learning:} Metric learning can be beneficial whenever we want to keep certain similarity relationship between samples \cite{bellet2013survey}. The Siamese network \cite{bromley1994signature} is a seminal work in deep metric learning, which minimizes the distance between a pair of samples with the same label and pushes samples with different labels apart. It was improved by the triplet network \cite{schroff2015facenet}, which considers relative distances. The triplet network requires three samples to compute a loss: an anchor $ A $, a positive sample $ P $ with the same label as $ A $, and a negative sample $ N $ with a different label. The network learns embeddings that respect the following distance relationship: 
\begin{equation}\label{eq:triplet}
\|f(A)-f(P)\|^2_2 + m < \|f(A)-f(N)\|^2_2,
\end{equation}
where $ f $ is the embedding function to be learned and $ m $ is a predefined margin. Various improvements to the standard triplet network have been proposed \cite{Zhang2016embedding, sohn16npair, Chen2017beyond, song17deep, Son2017multi}. Three key aspects in these methods are: how to define similarity between images, how to sample images for comparison, and how to compute the loss function.  Zhang et al.\ \cite{Zhang2016embedding} generalized the sampling strategy and triplet loss for multiple labels with hierarchical structures or shared attributes. Son et al.\ \cite{Son2017multi} employed label hierarchy to learn object embeddings for tracking, where object class is a high-level label and detection timestamp is low-level. Our sequential sampling strategy shares the similar spirit with them, but we lack well-defined supervision cues in the dataset, so we proposed strategies to leverage weak cues, \eg{} self-supervised body-part regressor and iterative refinement.



{\bf Lesion Management:} Great efforts have been devoted to lesion detection \cite{teramoto2016automated, Wang2017zoomin}, classification \cite{Cheng2016CAD, esteva2017dermatologist}, segmentation \cite{Cai2018MICCAI}, and retrieval \cite{Ramos2016content, Wei2016sim, tsochatzidis2017computer, Li2018review}. Recently, CNNs have become the method of choice over handcrafted features due to the former's superior performance \cite{Shin2016tmi, Tajbakhsh2016tmi, Greenspan2016guest, Litjens2017survey}. Our work is in line with content-based medical image retrieval, which has been surveyed in detail by \cite{Li2018review}. Existing methods generally focus on one type of lesion (\eg lung lesion or mammographic mass) and learn the similarity relationship based on manually annotated labels \cite{Wei2016sim, tsochatzidis2017computer} or radiology reports \cite{Ramos2016content}. To the best of our knowledge, no work has been done on learning deep lesion embeddings on a large comprehensive dataset with weak cues. Taking a different approach, \cite{hofmanninger2016unsupervised, wang2017unsupervised} cluster images or lesions to discover concepts in unlabeled large-scale datasets. However, they did not leverage multiple cues to explicitly model the semantic relationship between lesions. Several existing works on intra-patient lesion matching focus on detecting follow-up lesions and matching them pixel by pixel \cite{Hong2008auto, moltz2009general, silva2011fast, vivanti2015automatic}, which generally requires organ segmentation or time-consuming nonrigid volumetric registration. Besides, they are designed for certain types of lesions, whereas our lesion embedding can be used to match all kinds of lesions.

\section{DeepLesion Dataset}
\label{sec:ds}

The DeepLesion dataset\footnote{\url{https://nihcc.box.com/v/DeepLesion}} consists of over 32K clinically significant findings mined from a major institute's PACS. To the best of our knowledge, this dataset is the first to automatically extract lesions from challenging PACS sources. Importantly, the workflow described here can be readily scaled up and applied to multiple institutional PACS, providing a means for truly massive scales of data.

Radiologists routinely annotate clinically meaningful findings in medical images using arrows, lines, diameters or segmentations. These images, called ``bookmarked images'', have been collected over close to two decades in our institute's PACS. Without loss of generality, we study one type of bookmark in CT images: lesion diameters. As part of the RECIST guidelines \cite{Eisen09RECIST}, which is the standard in tracking lesion progression in the clinic, lesion diameters consist of two lines, one measuring the longest diameter and the second measuring its longest perpendicular diameter in the plane of measurement. We extract the lesion diameter coordinates from the PACS server and convert them into corresponding positions on the image plane. After removing some erroneous annotations, we obtain 32,120 axial CT slices (mostly 512 $\times$ 512) from 10,594 studies of 4,427 unique patients. There are 1--3 lesions in each image, adding up to 32,735 lesions altogether. We generate a box tightly around the two diameters and add a 5-pixel padding in each direction to capture the lesion's full spatial extent. Samples of the lesions and bounding boxes are in \Fig{dataset-vis}. More introduction of the dataset can be found in the supplementary material.

The 12-bit CT intensity range is rescaled to floating-point numbers in $ [0,255] $ using a single windowing covering the intensity ranges in lungs, soft tissues, and bones. Each image is resized so that the spacing is 1 mm/pixel. For each lesion, we crop a patch with 50 mm padding around its bounding box. To encode 3D information, we use 3 neighboring slices (interpolated at 2 mm slice intervals) to compose a 3-channel image. No data augmentation was used.

\section{Learning Lesion Embeddings}
\label{sec:lesEmb}
To learn lesion embeddings, we employ a triplet network with sequential sampling, as illustrated in \Fig{framework}. The cues used to supervise the network and the training strategy are described below.

\subsection{Supervision Cues}
\label{subsec:supInfo}

Supervision information, or cues, are key in defining the similarity relationship between lesions. Because it is prohibitively time-consuming to manually annotate all lesions in a PACS-based dataset like DeepLesion, a different approach must be employed. Here we use the cues of lesion type, relative body location, and size. {\bf Size information} (lengths of long and short lesion diameters) has been annotated by radiologists and ranges from 0.2 to 343 mm with a median of 15.6 mm. They are significant indicators of patients' conditions according to the RECIST guideline \cite{Eisen09RECIST}. For example, larger lymph nodes are considered lesions while those with short diameters $ < $ 10 mm are treated as normal \cite{Eisen09RECIST}. While size can be obtained directly from radiologists' markings, type and relative body location require more complex approaches.

{\bf Lesion Type:} Among all 32,735 lesions, we randomly select 30\% and manually label them into 8 types: lung, abdomen, mediastinum, liver, pelvis, soft tissue,  kidney, and bone. These are coarse-scale attributes of the lesions. An experienced radiologist verified the labels. The mediastinum class mainly consists of lymph nodes in the chest. Abdomen lesions are miscellaneous ones that are not in liver or kidney. The soft tissue class contains lesions in the muscle, skin, fat, etc. Among the labeled samples, we randomly select 25\% as training seeds to predict pseudo-labels, 25\% as the validation set, and the other 50\% as the test set. There is no patient-level overlap between all subsets.

The type of a lesion is related to its location, but the latter information cannot replace the former because some lesion types like bone and soft tissue have widespread locations. Neighboring types such as lung/mediastinum and abdomen/liver/kidney are hard to classify solely by location. The challenge with using PACS data is that there are no annotated class labels for each lesion in DeepLesion. Therefore, we use labeled seed samples to train a classifier and apply it to all unlabeled samples to get their pseudo-labels \cite{lee2013pseudo}. Details on the classifier are provided in Sec.\ \ref{subsec:network}.

{\bf Relative Body Location:} Relative body location is an important and clinically relevant cue in lesion characterization. While the $ x $ and $ y $ coordinates of a lesion are easy to acquire in axial CT slices, the $ z $ coordinate (\eg 0--1 from head to toe) is not as straightforward to find. The slice indices in the volume cannot be used to compute $ z $ because CT volumes often have different scan ranges (start, end), not to mention variabilities in body lengths and organ layouts. For this reason, we use the self-supervised body part regressor (SSBR), which provides a relative $z$ coordinate based on context appearance. 

\begin{figure}[h]
	\begin{center}
		\includegraphics[width=\linewidth,trim=0 0 0 0,clip]{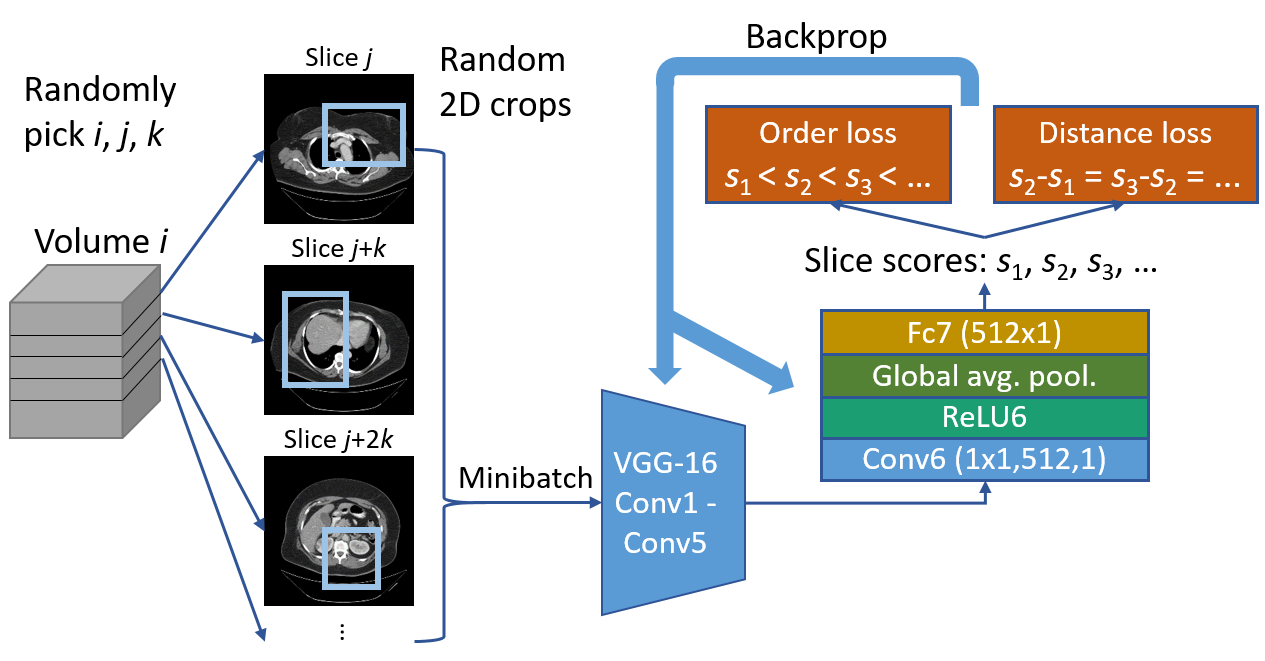}
	\end{center}
	\caption{Framework of the self-supervised body part regressor (SSBR).}
	\label{fig:ssbr}
\end{figure}

SSBR operates on the intuition that volumetric medical images are intrinsically structured, where the position and appearance of organs are relatively aligned. The superior-inferior slice order can be leveraged to learn an appearance-based $ z $. SSBR randomly picks $ m $ equidistant slices from a volume, denoted $ j, j+k, \ldots, j+k(m-1) $, where $ j $ and $ k $ are randomly determined. They are passed through a CNN to get a score $ s $ for each slice, which is optimized using the following loss function:
\begin{equation}\label{eq:SSBR}
\begin{aligned}
L_{\text{SSBR}} &=  L_{\text{order}} + L_{\text{dist}}; \\
L_{\text{order}}&=-\sum\nolimits_{i=0}^{m-2}{\log h\left(s_{j+k(i+1)}-s_{j+ki}\right)}; \\
L_{\text{dist}} &= \sum\nolimits_{i=0}^{m-3}{g(\Delta_{i+1}-\Delta_{i})}, \\
\Delta_{i} &=\, s_{j+k(i+1)}-s_{j+ki},
\end{aligned}
\end{equation}
where $ h $ is the sigmoid function, $ g $ is the smooth L1 loss \cite{Girshick2015fast}. $ L_{\text{order}} $ requires slices with larger indices to have larger scores. $ L_{\text{dist}} $ makes the difference between two slice scores proportional to their physical distance. The order loss and distance loss terms collaborate to push each slice score towards the correct direction relative to other slices. After convergence, slices scores are normalized to $ [0,1] $ to obtain the $ z $ coordinates without having to know which score corresponds to which body part. The framework of SSBR is shown in \Fig{ssbr}.

In DeepLesion, some CT volumes are zoomed in on a portion of the body, \eg{} only the left half is shown. To handle them, we train SSBR on random crops of the axial slices. Besides, SSBR does not perform well on body parts that are rare in the training set, \eg{} head and legs. Therefore, we train SSBR on all data first to detect hard volumes by examining the correlation coefficient ($ r $) between slice indices and slice scores, where lower $ r $ often indicates rare body parts in the volume. Then, SSBR is trained again on a resampled training set with hard volumes oversampled.

\subsection{Sequential Sampling}
\label{subsec:seqSmp}

Similar to \cite{Zhang2016embedding, Son2017multi}, we leverage multiple cues to describe the relationship between samples. A na\"{\i}ve strategy would be to treat all cues equally, where similarity can be calculated by, for instance, averaging the similarity of each cue. Another strategy assumes a hierarchical structure exists in the cues. Some high-level cues should be given higher priority. This strategy applies to our task, because intuitively lesions of the same type should be clustered together first. Within each type, we hope lesions that are closer in location to be closer in the feature space. If two lesions are similar in both type and location, they can be further ranked by size. This is a conditional ranking scheme.

\begin{figure}[tbp]
	\begin{center}
		\includegraphics[width=\linewidth]{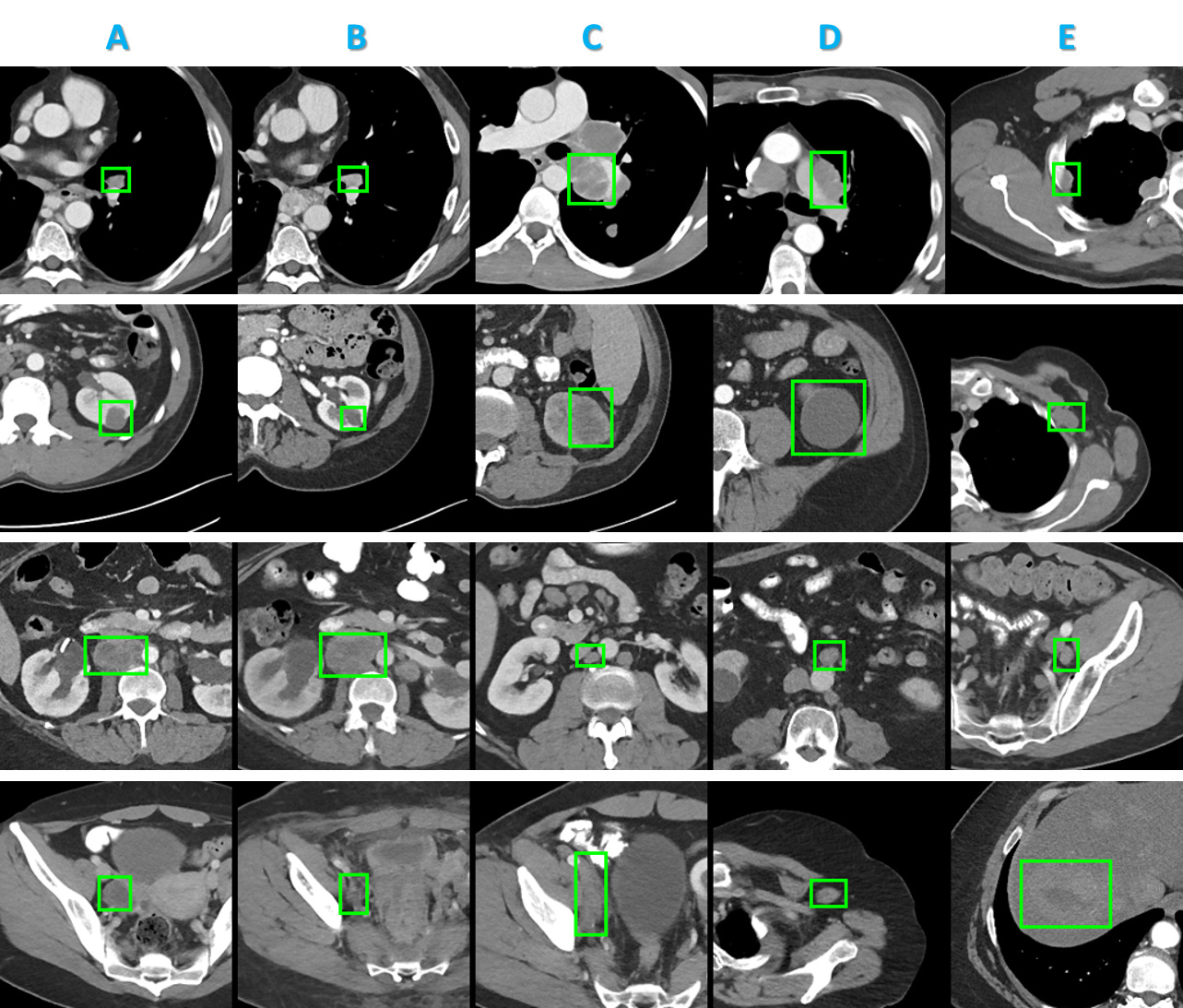}
	\end{center}
	\caption{Sample training sequences. Each row is a sequence. Columns 1--5 are examples of lesions $ A $--$ E $ in \Fig{framework}, respectively.}
	\label{fig:smpSeq}
\end{figure}

To this end, we adopt a sequential sampling strategy to select a sequence of lesions following the hierarchical relationship above. As depicted in \Fig{framework}, an anchor lesion $ A $ is randomly chosen first. Then, we look for lesions with similar type, location, and size with $ A $ and randomly pick $ B $ from the candidates. Likewise, $ C $ is a lesion with similar type and location but dissimilar size; $ D $ is similar in type but dissimilar in location (its size is not considered); $ E $ has a different type (its location and size are not considered). Here, two lesions are similar in type if they have the same pseudo-label; they are similar in location (size) if the Euclidean distance between their location (size) vectors is smaller than a threshold $ T_{\text{low}} $, whereas they are dissimilar if the distance is larger than $ T_{\text{high}} $. We do not use hard triplet mining as in \cite{schroff2015facenet, oh2016deep} because of the noisy cues. \Fig{smpSeq} presents some examples of lesion sequences. Note that there is label noise in the fourth row, where lesion $ D $ does not have the same type with $ A $ -- $ C $ (soft tissue versus pelvis).

A selected sequence can be decomposed into three triplets: $ABC$, $ACD$ and $ADE$. However, they are not equal, because we hope two lesions with dissimilar types to be farther apart than two with dissimilar locations, followed by size. Hence, we apply larger margins to higher-level triplets \cite{Zhang2016embedding, Chen2017beyond}. Our loss function is defined as:
\begin{align}\label{eq:tripletLoss}
L=\frac{1}{2S}\sum_{i=1}^{S} \big[&\max(0,d^2_{AB}-d^2_{AC}+m_1) \\ \notag
+&\max(0,d^2_{AC}-d^2_{AD}+m_2)  \\ \notag
+&\max(0,d^2_{AD}-d^2_{AE}+m_3) \big]. \notag
\end{align}
$m_3>m_2 > m_1 > 0$ are the hierarchical margins; $ S $ is the number of sequences in each mini-batch; $ d_{ij} $ is the Euclidean distance between two samples in the embedding space. The idea in sequential sampling resembles that of SSBR (\Eq{SSBR}): ranking a series of samples to make them self-organize and move to the right place in the feature space.

\subsection{Network Architecture and Training Strategy}
\label{subsec:network}

\begin{figure}[tbp]
	\begin{center}
		\includegraphics[width=\linewidth]{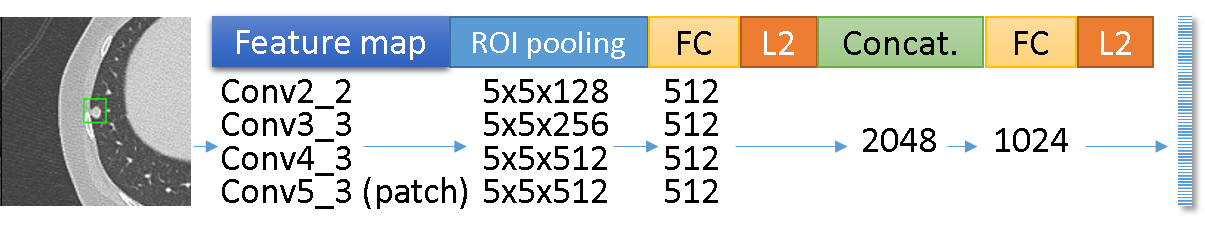}
	\end{center}
	\caption{Network architecture of the proposed triplet network.}
	\label{fig:network}
\end{figure}

VGG-16 \cite{Simonyan2015Vgg} is adopted as the backbone of the triplet network. As illustrated in \Fig{network}, we input the 50mm-padded lesion patch, then combine feature maps from 4 stages of VGG-16 to get a multi-scale feature representation with different padding sizes \cite{gidaris2015object, hu2017tiny}. Because of the variable sizes of the lesions, region of interest (ROI) pooling layers \cite{Girshick2015fast} are used to pool the feature maps to $5\times5\times num\_channel$ separately. For conv2\_2, conv3\_3, and conv4\_3, the ROI is the bounding box of the lesion in the patch to focus on its details. For conv5\_3, the ROI is the entire patch to capture the context of the lesion \cite{gidaris2015object, hu2017tiny}. Each pooled feature map is then passed through a fully-connected layer (FC), an L2 normalization layer (L2), and concatenated together. The final embedding is obtained after another round of FC and L2 normalization layers.

To get the initial embedding of each lesion, we use ImageNet \cite{Feifei09ImageNet} pretrained weights to initialize the convolutional layers, modify the output size of the ROI pooling layers to $ 1\times1\times num\_channel$, and remove the FC layers in \Fig{network} to get a 1408D feature vector. We use the labeled seed samples to train an 8-class RBF-kernel support vector machine (SVM) classifier and apply it to the unlabeled training samples to get their pseudo-labels. We also tried semi-supervised classification methods \cite{zhu2002learning, belkin2006manifold} and achieved comparable accuracy. Seed samples were not used to train the triplet network. We then sample sequences according to Sec.\ \ref{subsec:seqSmp} and train the triplet network until convergence. With the learned embeddings, we are able to retrain the initial classifier to get cleaner pseudo-labels, then fine-tune the triplet network with a lower learning rate \cite{wang2017unsupervised}. In our experiments, this iterative refinement improves performance.

\section{Lesion Organization}
\label{sec:lesionOrg}

The lesion graph can be constructed after the embeddings are learned. In this section, our two goals are content-based lesion retrieval and intra-patient lesion matching. The lesion graph can be used to directly tackle the first goal by finding nearest neighbors of query lesions. However, the latter one requires additional techniques to accomplish.

\subsection{Intra-patient Lesion Matching}
\label{subsec:lesionMatch}

We assume that lesions in all studies have been detected by other lesion detection algorithms \cite{teramoto2016automated} or marked by radiologists, which is the case in DeepLesion. In this section, our goal is to match the same lesion instances and group them for each patient. Potential challenges include appearance changes between studies due to lesion growth/shrinkage, movement of organs or measurement positions, and different contrast phases. Note that for one patient not all lesions occur in each study because the scan ranges vary and radiologists only mark a few target lesions. In addition, one CT study often contains multiple series (volumes) that are scanned at the same time point but differ in image filters, contrast phases, etc. To address these challenges, we design the lesion matching algorithm described in Algo.\ \ref{algo:matching}.

The basic idea is to build an intra-patient lesion graph and remove the edges connecting different lesion instances. The Euclidean distance of lesion embeddings is adopted as the similarity measurement. First, lesion instances from different series within the same study are merged if their distance is smaller than $ T_1 $. They are then treated as one lesion with embeddings averaged. Second, we consider lesions in all studies of the same patient. If the distance between two lesions is larger than $ T_2\, (T_2 > T_1)$, they are not similar and their edge is removed. After this step, one lesion in study 1 may still connect to multiple lesions in study 2 if they look similar, so we only keep the edge with the minimum distance and exclude the others. Finally, the remaining edges are used to extract the matched lesion groups.

\begin{algorithm}
	\caption{Intra-patient lesion matching}
	\begin{algorithmic}[1]
		\REQUIRE Lesions of the same patient represented by their embeddings; the study index $ s $ of each lesion; intra-study threshold $ T_1 $; inter-study threshold $ T_2 $.
		\ENSURE Matched lesion groups.
		\STATE Compute an intra-patient lesion graph $ G=\left<V,\mathcal{E}\right> $, where $ V $ are nodes (lesions) and $ \mathcal{E} $ are edges. Denote $ d_{ij} $ as the Euclidean distance between nodes $ i,j $.
		\STATE \textbf{Merge} nodes $ i$ and $j $ if $ s_i=s_j $ and $ d_{ij} < T_1 $.
		\STATE \textbf{Threshold}: $\mathcal{E} \gets \mathcal{E}-\mathcal{D}, \mathcal{D} = \{\left<i,j\right>\in\mathcal{E} | d_{ij} > T_2\}.~~$
		\STATE \textbf{Exclusion}: $\mathcal{E} \gets \mathcal{E}-\mathcal{C}, \mathcal{C} = \{\left<i,j\right> | \left<i,j\right>\in\mathcal{E},  \left<i,k\right>\in\mathcal{E}, s_j=s_k, \text{and } d_{ij}\geq d_{ik}\} $.
		\STATE \textbf{Extraction}: Each node group with edge connections is considered as a matched lesion group.
	\end{algorithmic}
	\label{algo:matching}
\end{algorithm}



\section{Experiments}
\label{sec:exp}

Our experiments aim to show that the learned lesion embeddings can be used to produce a semantically meaningful similarity graph for content-based lesion retrieval and intra-patient lesion matching. 

\subsection{Implementation Details}
\label{subsec:implement}

We empirically choose the hierarchical margins in \Eq{tripletLoss} to be $ m_1=0.1,m_2=0.2,m_3=0.4 $. The maximum value of each dimension of the locations and sizes is normalized to 1. When selecting sequences for training, the similarity thresholds for location and size are $T_{\text{low}} = 0.02, T_{\text{high}}=0.1$. We use $ S=24 $ sequences per mini-batch. The network is trained using stochastic gradient descent (SGD) with a learning rate of 0.002, which is reduced to 0.0002 in iteration 2K. After convergence (generally in 3K iterations), we do iterative refinement by updating the pseudo-labels and fine-tuning the network with a learning rate of 0.0002. This refinement is performed only once because we find that more iterations only add marginal accuracy improvements. For lesion matching, the intra-study threshold $ T_1$ is 0.1 and we vary the inter-study threshold $ T_2 $ to compute the precision-recall curve. Due to space limits, the details of SSBR are given in the supplementary material.

\subsection{Content-based Lesion Retrieval}
\label{subsec:exp:lesionEmb}

\begin{figure*}[tbp]
	\begin{center}
		\includegraphics[width=\linewidth]{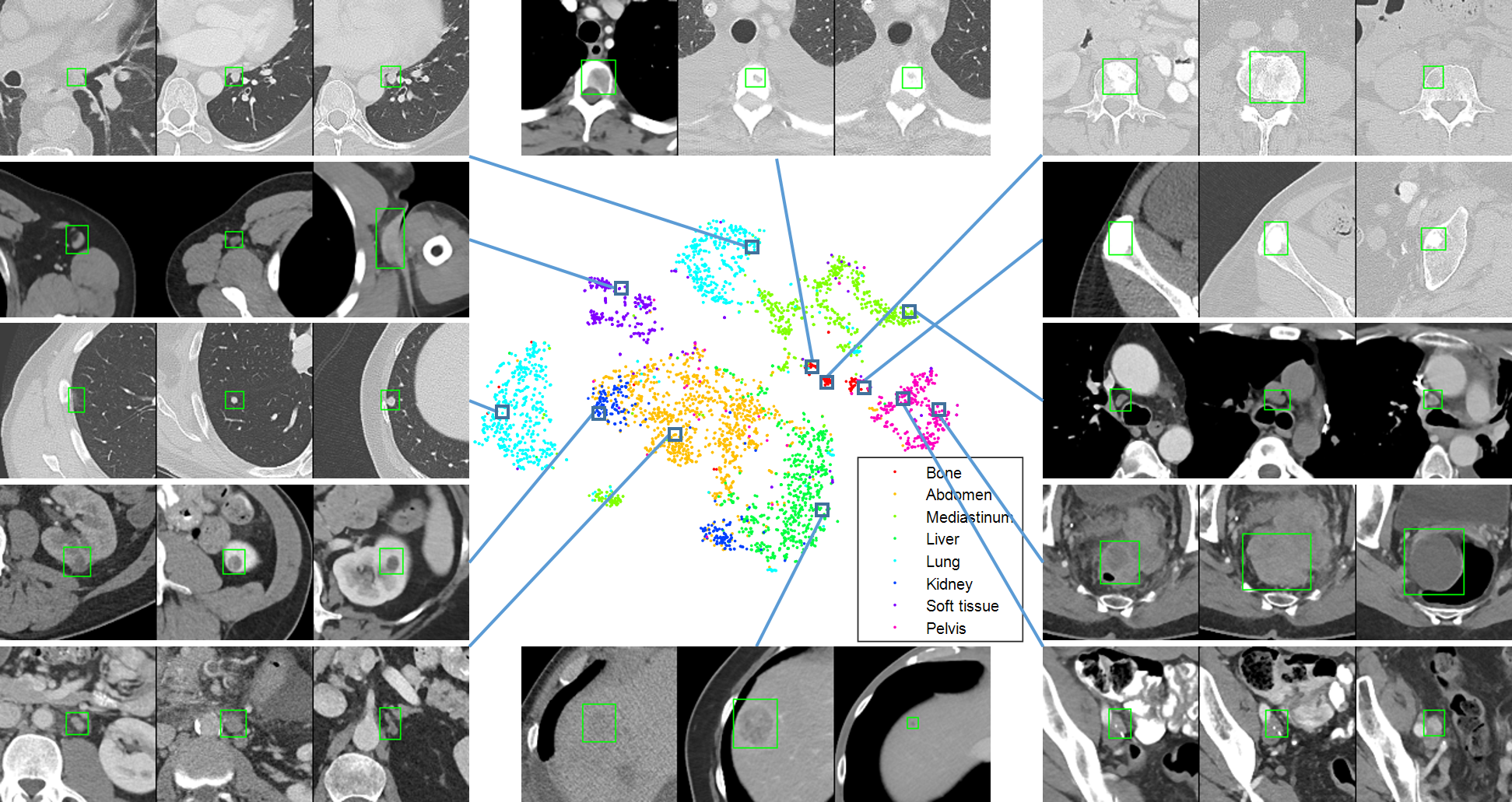}
	\end{center}
	\caption{t-SNE visualization of the lesion embeddings on the test set (4,927 samples) of DeepLesion. Colors indicate the manually labeled lesion types. We also split the samples to 128 clusters using K-means and show 3 random lesions in 12 representative clusters. We did not choose to show closest samples because they are very similar. Best viewed in color.}
	\label{fig:embedding}
\end{figure*}

First, we qualitatively investigate the learned lesion embeddings in \Fig{embedding}, which shows the Barnes-Hut t-SNE visualization \cite{Maaten2014tsne} of the 1024D embedding and some sample lesions. The visualization is applied to our manually labeled test set, where we have lesion-type ground truth. As we can see, there is a clear correlation between data clusters and lesion types. It is interesting to find that some types are split into several clusters. For example, lung lesions are separated to left lung and right lung, and so are kidney lesions. Bone lesions are split into 3 small clusters, which are found to be mainly chest, abdomen, and pelvis ones, respectively. Abdomen, liver, and kidney lesions are close both in real-world location and in the feature space. These observations demonstrate the embeddings are organized by both type and location. The sample lesions in \Fig{embedding} are roughly similar in type, location, and size. 

\begin{figure*}[]
	\begin{center}
		\includegraphics[width=\linewidth]{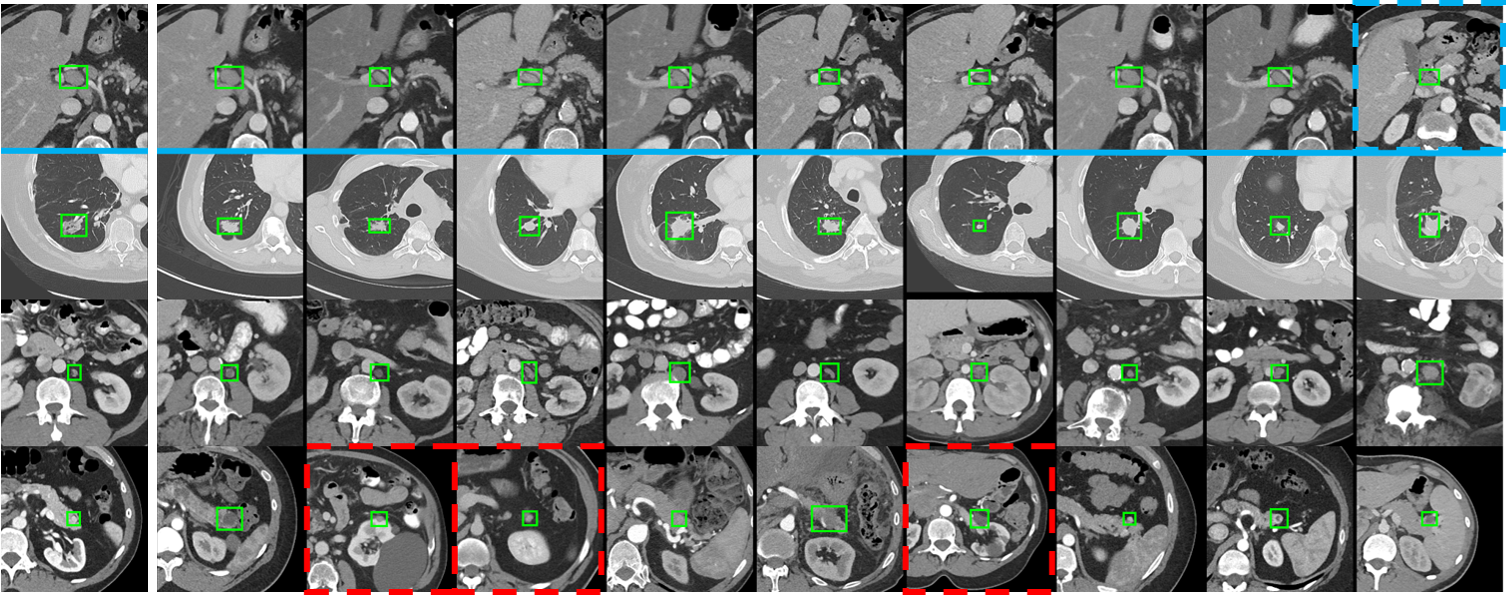}
	\end{center}
	\caption{Examples of query lesions (first column) and the top-9 retrieved lesions on the test set of DeepLesion. In the first row, the blue dashed box marks the lesion from a different patient than the query one, whereas the other 9 are all from the same patient. In rows 2--4, we constrain that the query and all retrieved lesions must come from different patients. Red dashed boxes indicate incorrect results, see text.}
	\label{fig:retrieval_samples}
\end{figure*}

\Fig{retrieval_samples} displays several retrieval results using the lesion embeddings. They are ranked by their Euclidean distance with the query one. We find that the top results are mostly the same lesion instances of the same patient, as shown in the first row of \Fig{retrieval_samples}. It suggests the potential of the proposed embedding on lesion matching, which will be further evaluated in the following section. To better exhibit the ability of the embedding in finding semantically similar lesions, rows 2--4 of \Fig{retrieval_samples} depict retrieved lesions from different patients. Spiculated nodules in the right lung and left para-aortic lymph nodes are retrieved in rows 2 and 3, respectively. Row 4 depicts lesions located on the tail of the pancreas, and also some failure cases marked in red. Note that our type labels used in supervision are too coarse to describe either abdomen lymph nodes or pancreas lesions (both are covered in the abdomen class). However, the framework naturally clusters lesions from the same body structures together due to similarity in type, location, size, and appearance, thus discovering these sub-types. Although appearance is not used as supervision information, it is intrinsically considered by the CNN-based feature extraction architecture and strengthened by the multi-scale strategy. To explicitly distinguish sub-types and enhance the semantic information in the embeddings, we can either enrich the type labels by mining knowledge from radiology reports \cite{shin2016interleaved, cornegruta2016modeling, Wang2017ChestXray8, zhang2017mdnet}, or integrate training samples from other medical image datasets with more specialized annotations \cite{Clark2013tcia, setio2017validation}. These new labels may be incomplete or noisy, which fits the setting of our system.

Quantitative experimental results on lesion retrieval, clustering, and classification are listed in \Tbl{embedding_res}. For retrieval, the three supervision cues are thoroughly inspected. Because location and size (all normalized to 0--1) are continuous labels, we define an evaluation criterion called average retrieval error (ARE):
\begin{equation}\label{eq:retrival_err}
\text{ARE}=\frac{1}{K}\sum_{i=1}^{K}\|\mathbf{t}^Q - \mathbf{t}^R_i\|_2,
\end{equation}
where $ \mathbf{t}^Q $ is the location or size of the query lesion and $ \mathbf{t}^R_i $ is that of the $ i $th retrieved lesion among the top-$ K $. On the other hand, the ARE of lesion type is simply $1- precision$. Clustering and classification accuracy are evaluated only on lesion type. Purity and normalized mutual information (NMI) of clustering are defined in \cite{Manning2008intro}. The multi-scale ImageNet feature is computed by replacing the 5$ \times $5 ROI pooling to 1$ \times $1 and removing the FC layers.

\begin{table*}
	\begin{center}
		\begin{tabular}{l|ccc|cc|c}
			\hline
			\multirow{2}{*}{Feature representation} & \multicolumn{3}{|c|}{Average retrieval error} & \multicolumn{2}{|c|}{Clustering} & \multicolumn{1}{|c}{Classification} \\
			\cline{2-7}
			& Type & Location & Size & Purity & NMI & Accuracy \\
			\hline\hline
			Baseline: Multi-scale ImageNet feature	& 15.2	& 9.6	& 6.9	& 58.7	& 35.8	& 86.2 \\
			Baseline: Location feature	& 22.4	& \textbf{2.5}	& 8.8	& 51.6	& 32.6	& 59.7 \\
			\hline
			Triplet with type	& ~~8.8$\pm$0.2	& 10.8$\pm$0.2	& 5.7$\pm$0.1	& 84.7$\pm$2.8	& 71.5$\pm$1.7	& 89.5$\pm$0.3	\\
			Triplet with location	& 13.0$\pm$0.1	& ~~6.5$\pm$0.1	& 6.2$\pm$0.1	& 61.1$\pm$4.4	& 39.5$\pm$4.3	& 87.8$\pm$0.5	\\
			Triplet with type + location	& ~~8.7$\pm$0.2	& ~~7.2$\pm$0.1	& 6.0$\pm$0.1	& 81.3$\pm$4.7	& 68.0$\pm$2.4	& 89.9$\pm$0.3	 \\ 
			Triplet with type + location + size	& ~~\textbf{8.5}$\pm$\textbf{0.1}	& ~~7.2$\pm$0.0	& \textbf{5.1}$\pm$\textbf{0.0}	& \textbf{86.0}$\pm$\textbf{3.9}	& \textbf{72.4}$\pm$\textbf{4.6}	& \textbf{90.5}$\pm$\textbf{0.2}	 \\
			\hline
			w/o Multi-scale feature: conv5	& 11.5$\pm$0.2	& ~~7.1$\pm$0.1	& 6.3$\pm$0.0	& 79.8$\pm$0.6	& 64.8$\pm$1.2	& 86.6$\pm$0.4	\\
			w/ ~~Multi-scale feature: conv5 + conv4	& ~~9.7$\pm$0.2	& ~~7.0$\pm$0.0	& 5.4$\pm$0.1	& 82.4$\pm$3.3	& 67.9$\pm$2.2	& 89.0$\pm$0.6 \\
			w/o Iterative refinement	& ~~8.7$\pm$0.2	& ~~7.3$\pm$0.0	& 5.2$\pm$0.1	& 85.4$\pm$2.8	& 69.8$\pm$2.0	& 90.2$\pm$0.2	 \\
			\hline
		\end{tabular}
	\end{center}
	\caption{Evaluation results on the test set (4,927 samples) of DeepLesion. For retrieval, we compute the average retrieval error (\%) in type, location, and size of the top-5 retrieved lesions compared to the query one. For clustering, lesions are clustered to 8 groups using K-means to calculate the purity and NMI (\%). For classification, we train a 8-way softmax classifier on the seed labeled samples and apply it on the test set. The CNN in each method was trained 5 times using different random seeds. Mean results and standard deviations are reported.}
	\label{tbl:embedding_res}
\end{table*}

\begin{figure}[]
	\begin{center}
		\includegraphics[width=.9\linewidth]{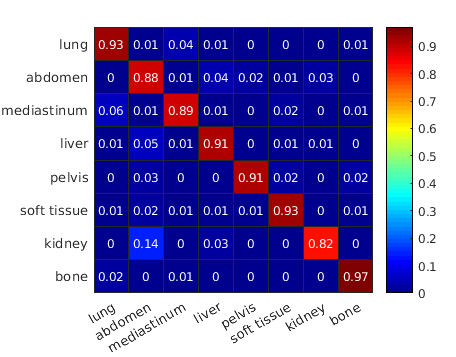}
	\end{center}
	\caption{The confusion matrix of lesion classification.}
	\label{fig:conf}
\end{figure}

In \Tbl{embedding_res}, the middle part compares the results of applying different supervision information to train the triplet network. Importantly, when location and size are added as supervision cues, our network performs best on lesion-type retrieval---even better than when only lesion-type is used as the cue. This indicates that location and size provides important supplementary information in learning similarity embeddings, possibly making the embeddings more organized and acting as regularizers. The bottom part of the table shows results of ablation studies, which demonstrate the effectiveness of multi-scale features and iterative refinement, highlighting the importance of combining visual features from different context levels. When only coarse-scale features (conv5, conv4) are used, location ARE is slightly better because location mainly relies on high-level context information. However, fusing fine-level features (conv3, conv2) significantly improves type and size prediction, which indicates that details are important in these aspects. We also inspected the confusion matrix for lesion classification (\Fig{conf}). The most confusing types are mediastinum/lung lesions, and abdomen/liver/kidney lesions, since some of them are similar in both appearance and location. More visual results are presented in the supplementary material.

\subsection{Intra-patient Lesion Matching}
\label{subsec:exp:lesionMatch}

\begin{figure}[]
	\begin{center}
		\includegraphics[width=\linewidth,trim=10 0 20 0,clip]{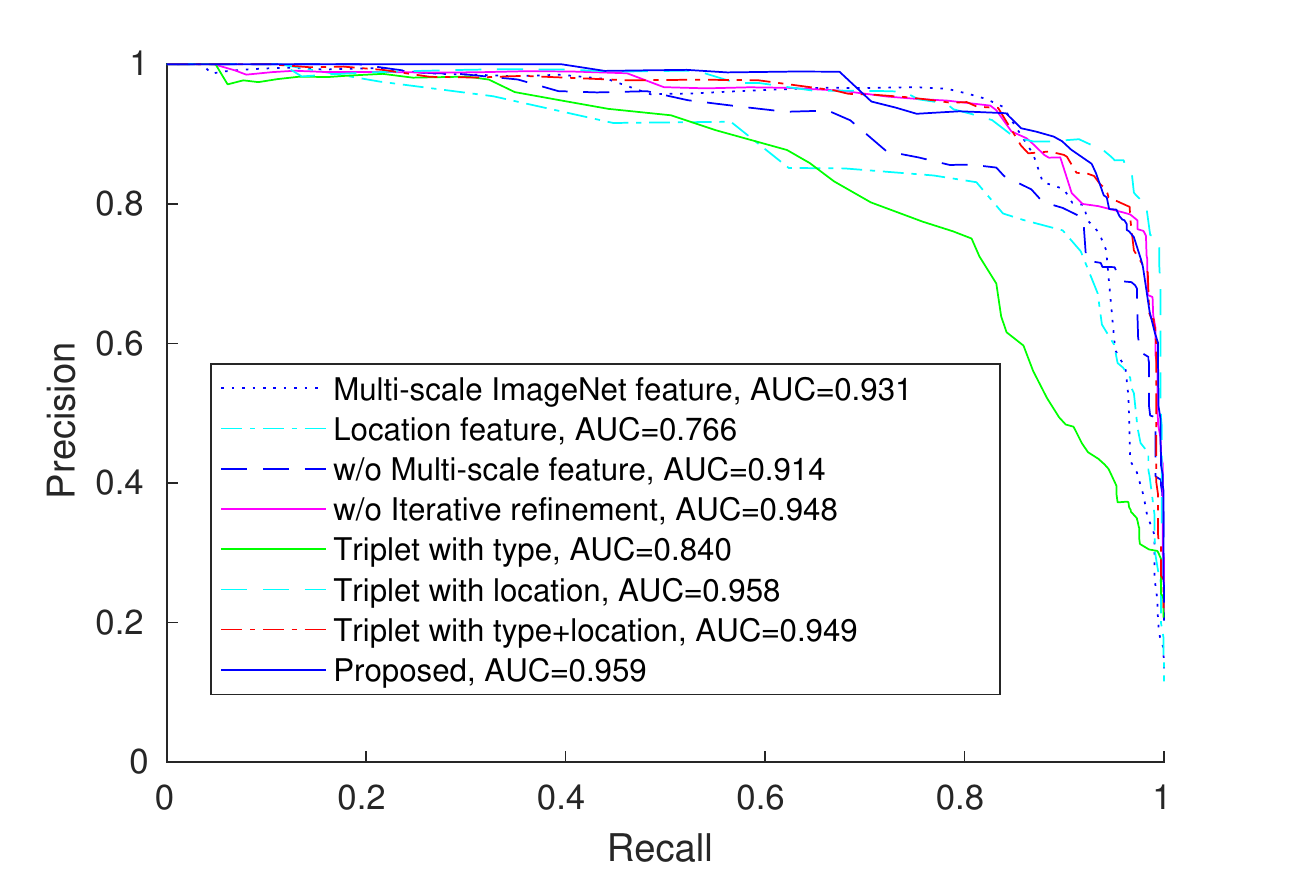}
	\end{center}
	\caption{Precision-recall curves of various methods on the intra-patient lesion matching task using DeepLesion. The area-under-curve (AUC) values are shown in the legends.}
	\label{fig:matching_PR}
\end{figure}

We manually grouped 1313 lesions from 103 patients in DeepLesion to 593 groups for evaluation. Each group contains instances of the same lesion across time. There are 1--11 lesions per group. Precision and recall are defined according to \cite{Manning2008intro}, where a true positive decision assigns two lesions of the same instance to the same group, and a false positive decision assigns two lesions of different instances to the same group, etc. As presented in \Fig{matching_PR}, our proposed embedding achieves the highest area under the curve (AUC). The location feature does not perform well because different lesion instances may be close to each other in location. This problem can be mitigated by combining location with appearance and using multi-scale features (accomplished in our triplet network). Our algorithm does not require any annotation of matched lesions for training. It is appearance-based and needs no registration or organ mask, thus is fast.

\section{Conclusion and Future Work}
\label{sec:conclusion}

In this paper, we present a large-scale and comprehensive dataset, DeepLesion, which contains significant radiology image findings mined from PACS. Lesion embeddings are learned with a triplet network to model their similarity relationship in type, location, and size. The only manual efforts needed are the class labels of some seed images. Promising results are obtained in content-based lesion retrieval and intra-patient lesion matching. The framework can be used as a generic lesion search engine, classifier, and matching tool. After being classified or retrieved by our system, lesions can be further processed by other specialist systems trained on data of a certain type. In the future, we plan to incorporate more fine-grained semantic information (\eg from radiology reports, other specialized datasets, or active learning) to build a lesion knowledge graph.

\section{Supplementary Material}

In this section, we provide some additional illustrations of the paper. \Sec{dataset} visualizes the DeepLesion dataset and describes some details. \Sec{SSBR} provides implementation details of the self-supervised body-part regressor. More content-based lesion retrieval results are presented in \Sec{retrieval}. \Sec{match} illustrates the intra-patient lesion matching task and the intra-patient lesion graph.

\subsection{DeepLesion Dataset: Visualization and Details}
\label{sec:dataset}

\begin{figure}[]
	\begin{center}
		\includegraphics[width=\linewidth,trim=150 120 420 100,clip]{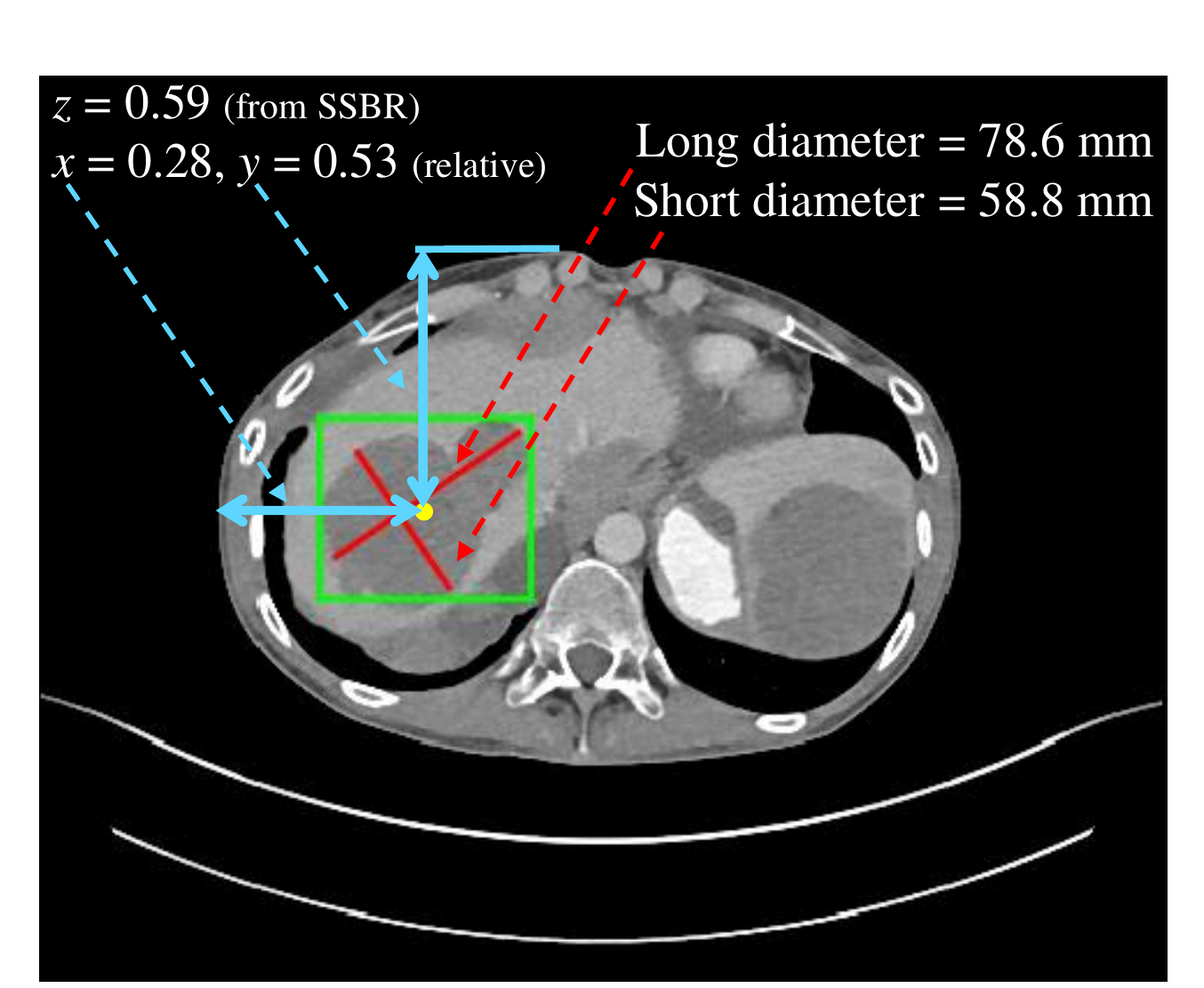}
	\end{center}
	\caption{Location and size of a sample lesion. The red lines are the long and short diameters annotated by radiologists during their daily work. The green box is the bounding box calculated from the diameters. The yellow dot is the center of the bounding box. The blue lines indicate the relative $ x $- and $ y $-coordinates of the lesion. The $ z$-coordinate is predicted by SSBR. Best viewed in color.}
	\label{fig:example-measure}
\end{figure}

\begin{figure}[]
	\begin{center}
		\includegraphics[width=\linewidth,trim=0 20 20 10,clip]{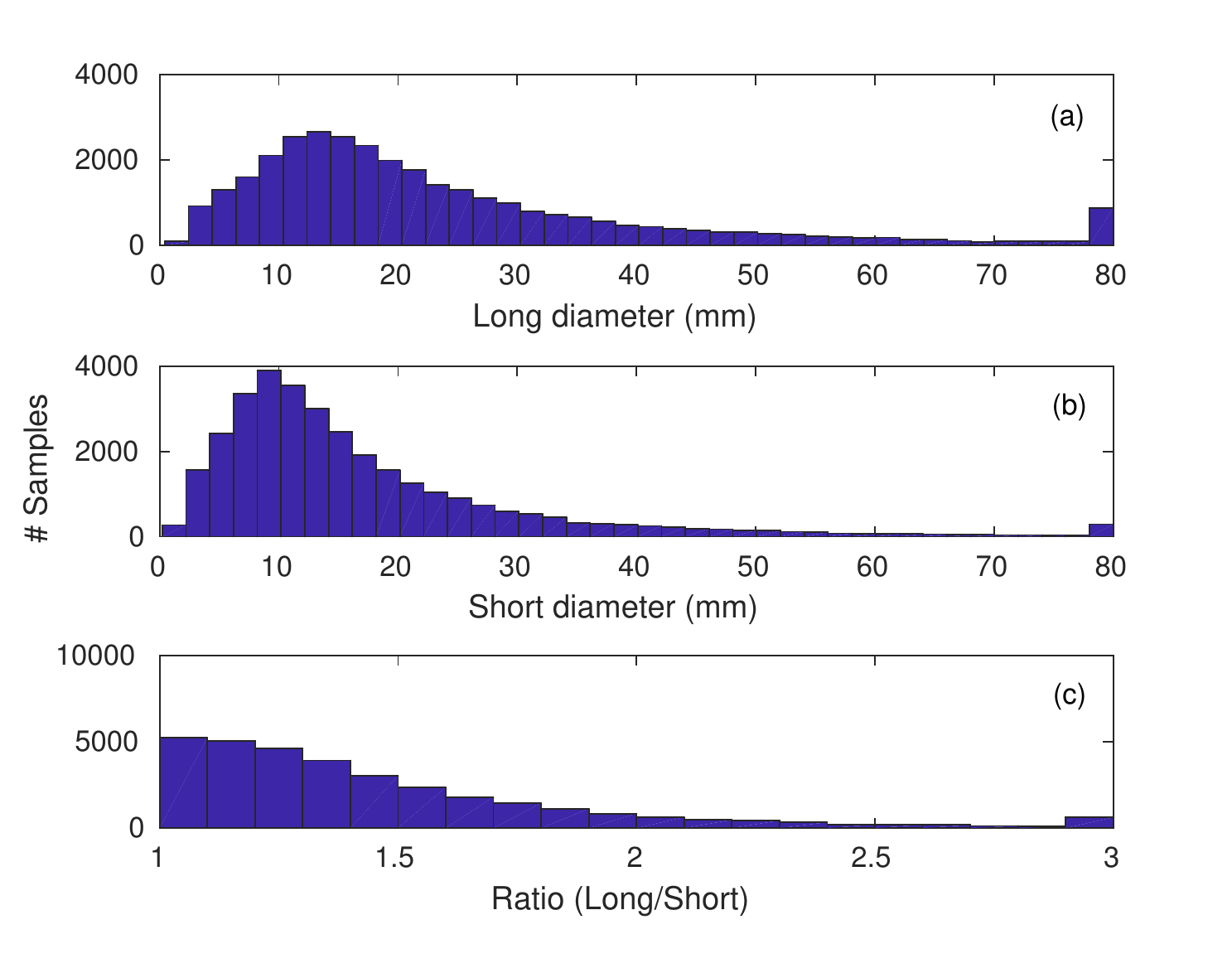}
	\end{center}
	\caption{Distribution of the lesion-sizes in DeepLesion. For clarity, values greater than the upper bound of the $ x $-axis of each plot are grouped in the last bin of each histogram.}
	\label{fig:size-distrib}
\end{figure}

\begin{figure*}[]
	\begin{center}
		\includegraphics[width=.9\linewidth,trim=10 60 0 50,clip]{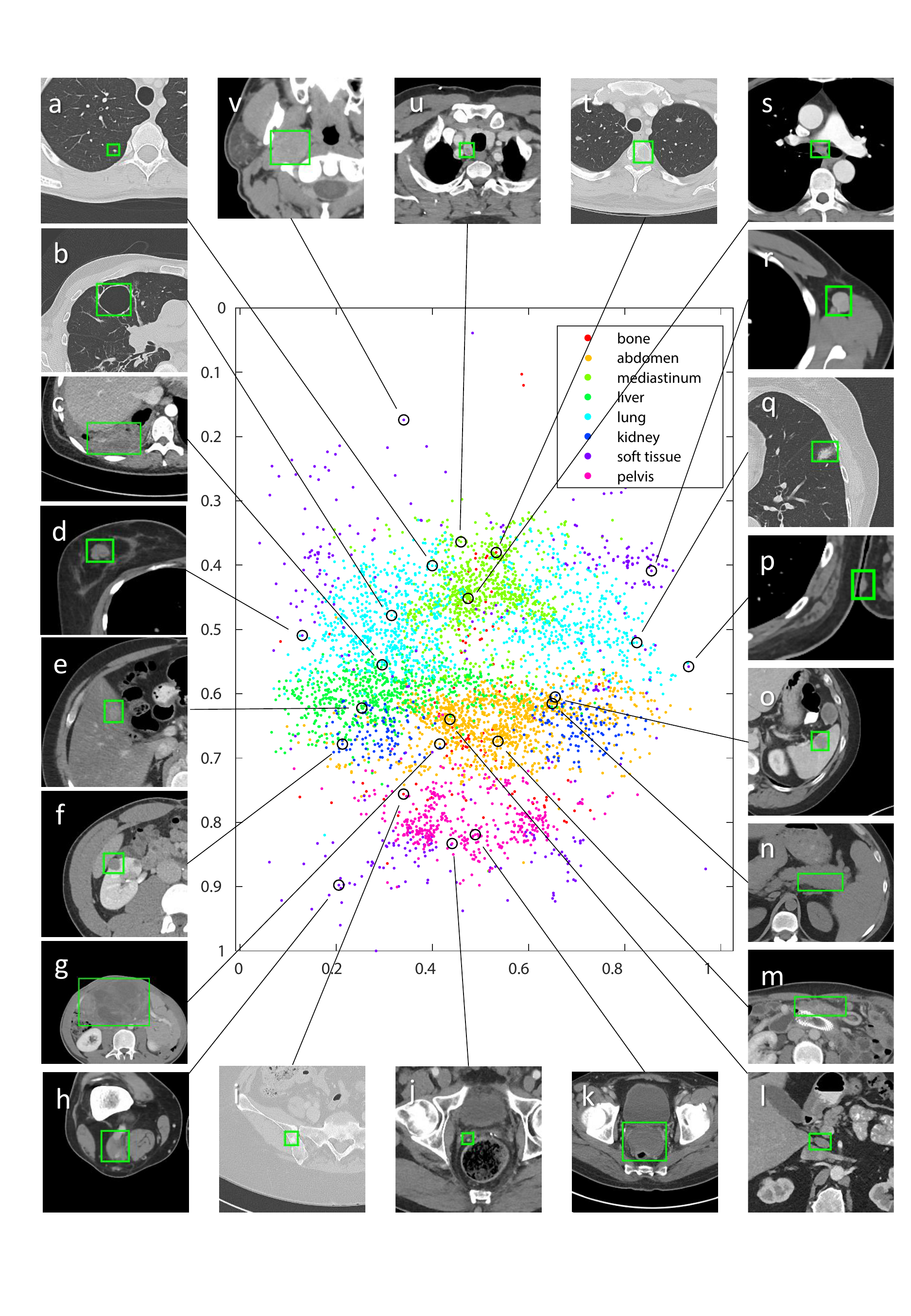}
	\end{center}
	\caption{Visualization of the DeepLesion dataset (test set). The $ x$- and $ y $-axes of the scatter map correspond to the $ x $- and $ z $-coordinates of the relative body location of each lesion, respectively. Therefore, this map is similar to a frontal view of the human body. Colors indicate the manually labeled lesion types. Sample lesions are exhibited to show the great diversity of DeepLesion, including: a.\ lung nodule; b.\ lung cyst; c.\ costophrenic sulcus (lung) mass/fluid; d.\ breast mass; e.\ liver lesion; f.\ renal mass; g.\ large abdominal mass; h.\ posterior thigh mass; i.\ iliac sclerotic lesion; j.\ perirectal lymph node (LN); k.\ pelvic mass; l.\ periportal LN; m.\ omental mass; n.\ peripancreatic lesion; o.\ splenic lesion; p.\ subcutaneous/skin nodule; q.\ ground glass opacity; r.\ axillary LN; s.\ subcarinal LN; t.\ vertebral body metastasis; u.\ thyroid nodule; v.\ neck mass.}
	\label{fig:dataset-vis}
\end{figure*}

To provide an overview of the DeepLesion dataset, we draw a scatter map to show the distribution of the types and relative body locations of the lesions in \Fig{dataset-vis}. From the lesion types and sample images, one can see that the relative body locations of the lesions are consistent with their actual physical positions, proving the validity of the location information used in the paper, particularly the self-supervised body-part regressor. Some lesion types like bone and soft tissue have widespread locations. Neighboring types such as lung/mediastinum and abdomen/liver/kidney have large overlap in location due to inter-subject variabilities. Besides, we can clearly see the considerable diversity of DeepLesion.

\Fig{example-measure} illustrates the approach to obtain the location and size of a lesion. In order to locate a lesion in the body, we first obtain the mask of the body in the axial slice, then compute the relative position (0--1) of the lesion center to get the $ x $- and $ y $-coordinates. As for $ z $, the self-supervised body-part regressor (SSBR) is used. We also show the distribution of the lesion-sizes in \Fig{size-distrib}.

\subsection{Self-Supervised Body-Part Regressor: Implementation Details}
\label{sec:SSBR}

To train SSBR, we randomly pick 800 unlabeled CT volumes of 420 subjects from DeepLesion. Each axial slice in the volumes is resized to 128 $\times$ 128 pixels. No further preprocessing or data augmentation was performed. In each mini-batch, we randomly select 256 slices from 32 volumes (8 equidistant slices in each volume, see Eq.\ 2 in the paper) for training. The network is trained using stochastic gradient descent with a learning rate of 0.002. It generally converges in 1.5K iterations.

The sample lesions in \Fig{dataset-vis} can be used to qualitatively evaluate the learned slice scores, or $ z $-coordinates. We also conducted a preliminary experiment to quantitatively assess SSBR. A test set including 18,195 slices subsampled from 260 volumes of 140 new subjects are collected. They are manually labeled as one of the 3 classes: chest (5903 slices), abdomen (6744), or pelvis (5548). The abdomen class starts from the upper border of the liver and ends at the upper border of the ilium. Then, we set two thresholds on the slice scores to classify them to the three classes. The classification accuracy is 95.99\%, with all classification errors appearing at transition regions (chest-abdomen, abdomen-pelvis) partially because of their ambiguity. The result proves the effectiveness of SSBR. More importantly, SSBR is trained on unlabeled volumes that are abundant in every hospital's database, thus zero annotation effort is needed.


\subsection{Content-based Lesion Retrieval: More Results}
\label{sec:retrieval}

\begin{figure*}[]
	\begin{center}
		\includegraphics[width=1\linewidth,trim=0 180 10 60,clip]{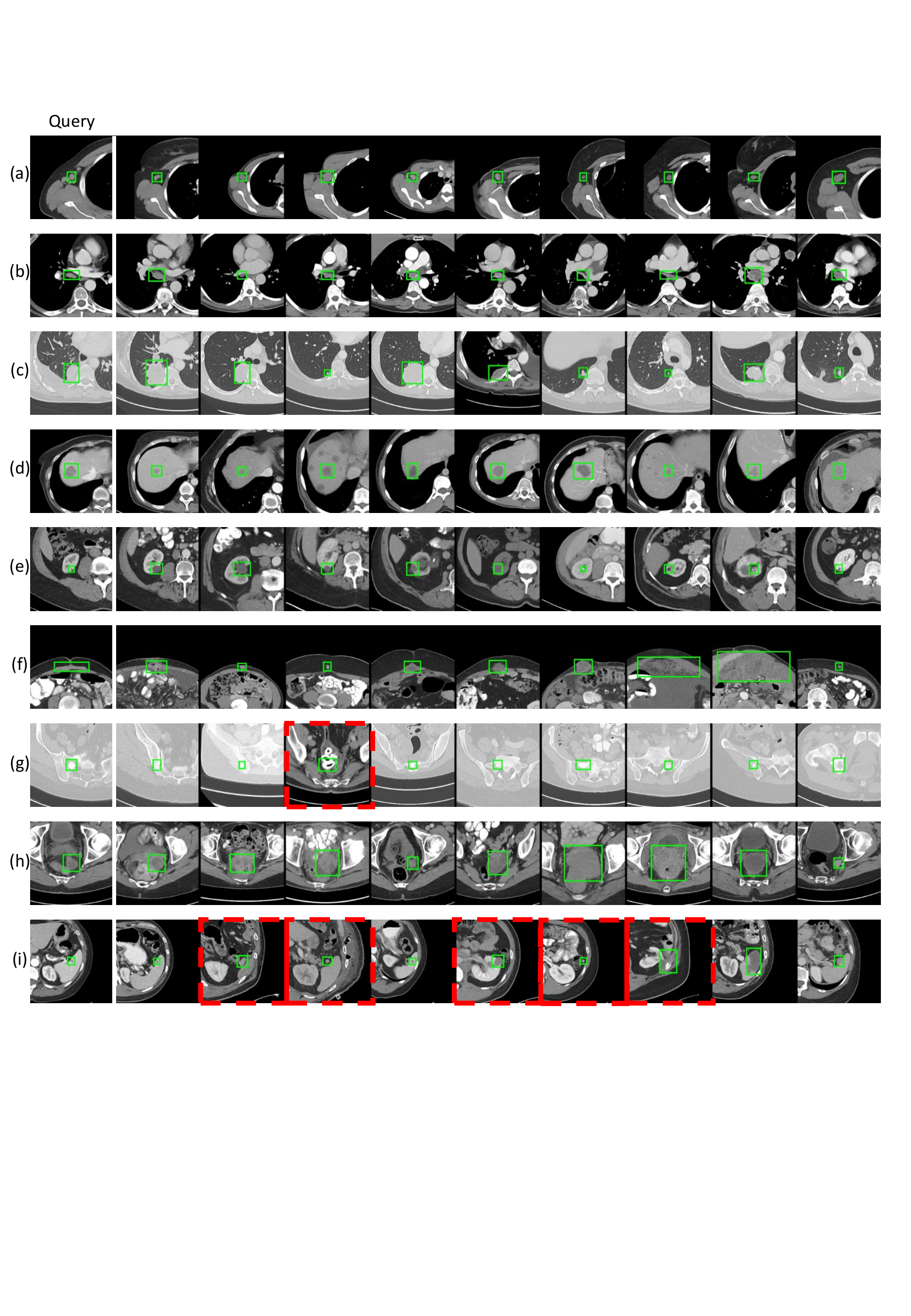}
	\end{center}
	\caption{More examples of query lesions (first column) and the top-9 retrieved lesions on the test set of DeepLesion. We constrain that the query and all retrieved lesions must come from different patients. Red dashed boxes indicate incorrect results. The lesions in each row are: (a) Right axillary lymph nodes; (b) subcarinal lymph nodes; (c) lung masses or nodules near the pleura; (d) liver lesions near the liver dome; (e) right kidney lesions; (f) lesions near the anterior abdomen wall; (g) lesions on pelvic bones except the one in the red box, which is a peripherally calcified mass. (h) inferior pelvic lesions; (i) spleen lesions except the ones in red boxes.}
	\label{fig:r2}
\end{figure*}

More examples of lesion retrieval are shown in \Fig{r2}. We try to exhibit typical examples of all lesion types. The last row is a failure case. Most retrieved lesions are similar with the query ones in type, location, and size. More importantly, most retrieved lesions and the query ones come from semantically similar body structures that are not specified in the training labels. The failure cases in \Fig{r2} have dissimilar types with the query ones. They were retrieved mainly because they have similar location, size, and appearance with query ones.

\subsection{Intra-Patient Lesion Matching: An Example}
\label{sec:match}

\begin{figure*}[]
	\begin{center}
		\includegraphics[width=\linewidth,trim=60 20 40 0,clip]{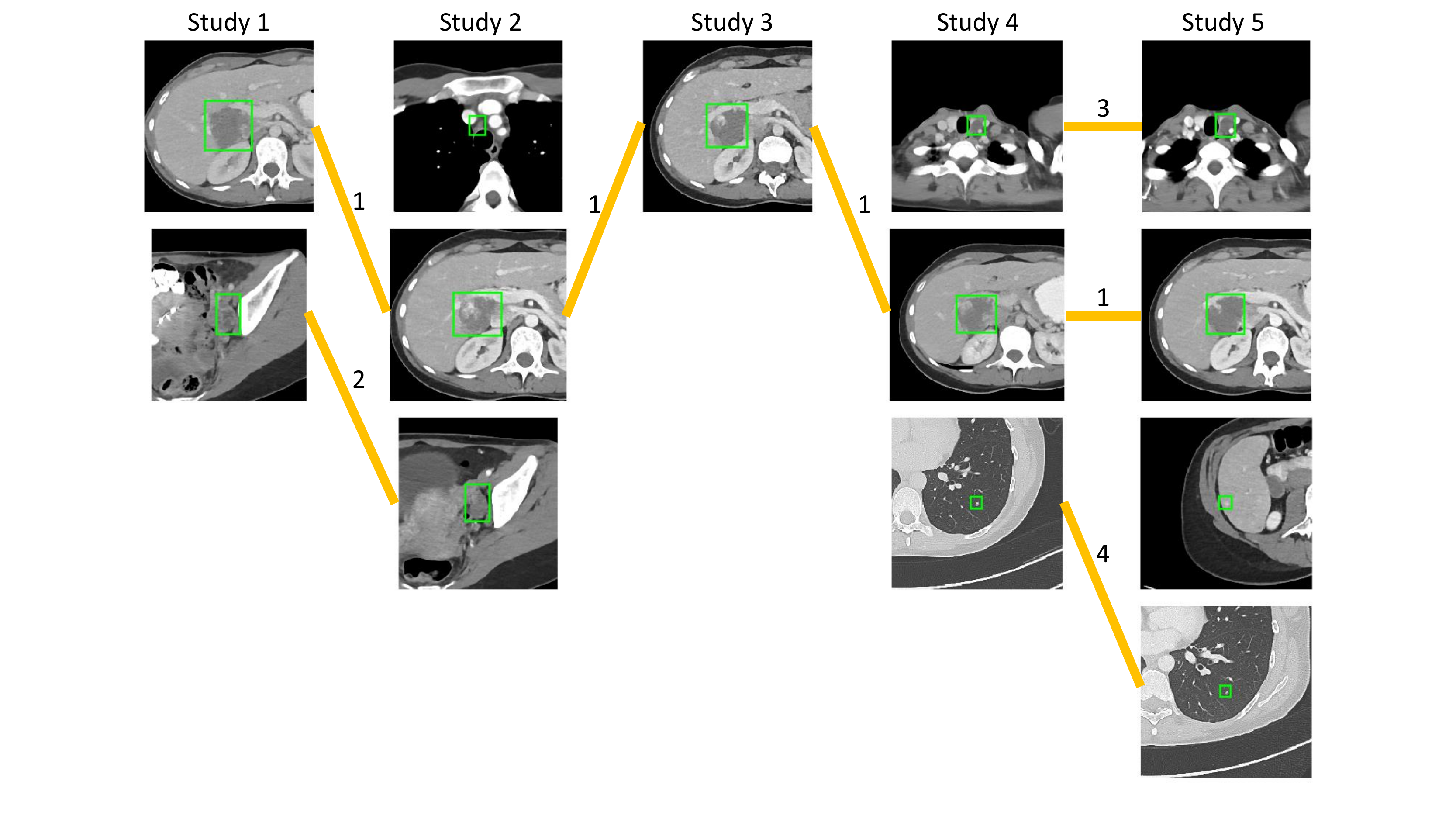}
	\end{center}
	\caption{All lesions of a sample patient in DeepLesion. Lesions in each study (CT examination) are listed in a column. Not all lesions occur in each study, because the scan ranges of each study vary and radiologists only mark a few target lesions. We group the same lesion instances to sequences. Four sequences are found and marked in the figure, where the numbers on the connections represent the lesion IDs.}
	\label{fig:matching-seq}
\end{figure*}

\begin{figure}[]
	\begin{center}
		\includegraphics[width=\linewidth,trim=80 60 60 40,clip]{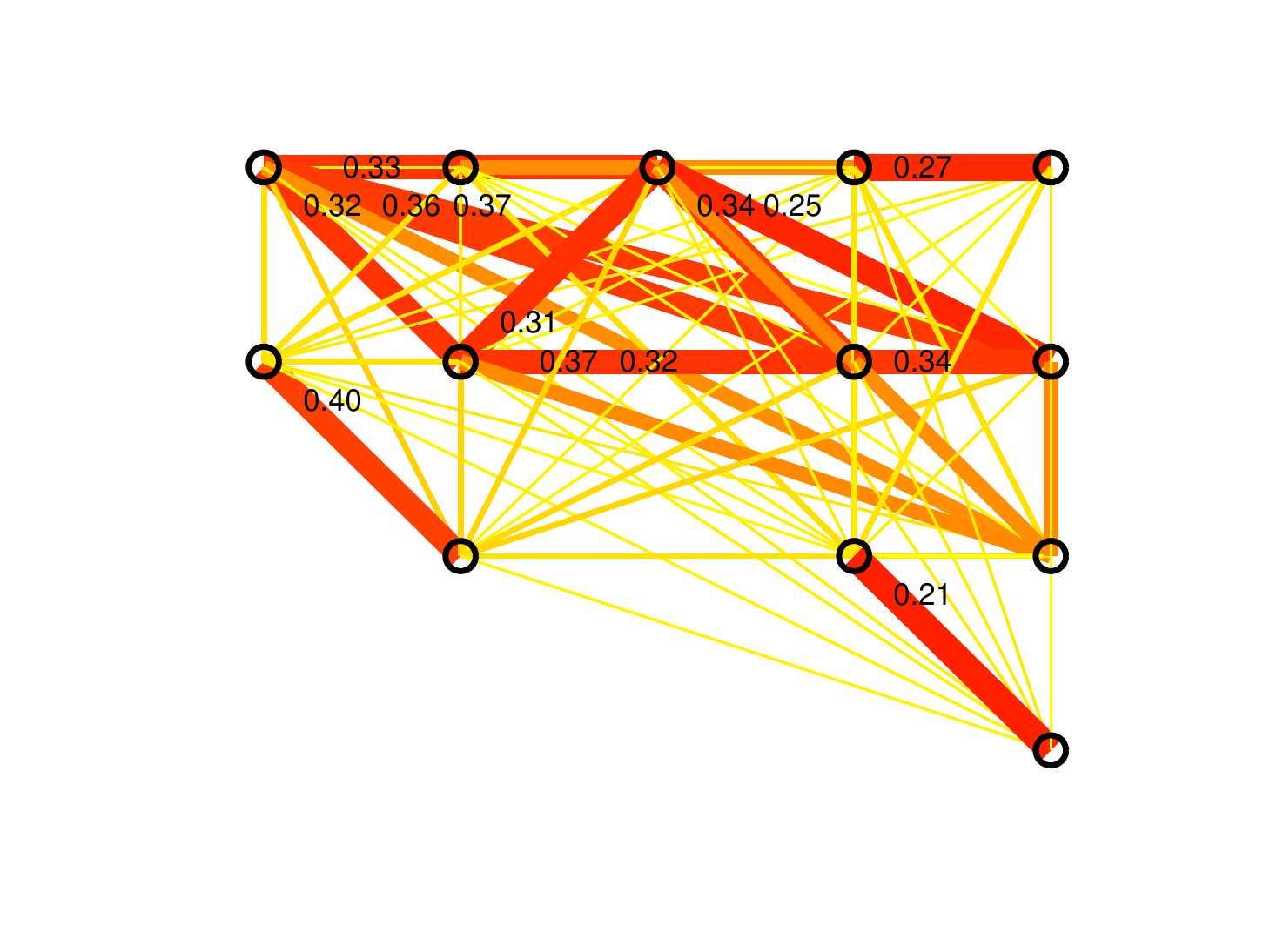}
	\end{center}
	\caption{The intra-patient lesion graph of the patient in \Fig{matching-seq}. For clarity, the lesions in \Fig{matching-seq} are replaced by nodes in this figure. The numbers on the edges are the Euclidean distances between nodes. We only show small distances in the figure. Red, thick edges indicate smaller distances. Note that some edges may overlap with other edges or nodes.}
	\label{fig:matching-complete}
\end{figure}

\begin{figure}[]
	\begin{center}
		\includegraphics[width=\linewidth,trim=80 60 60 40,clip]{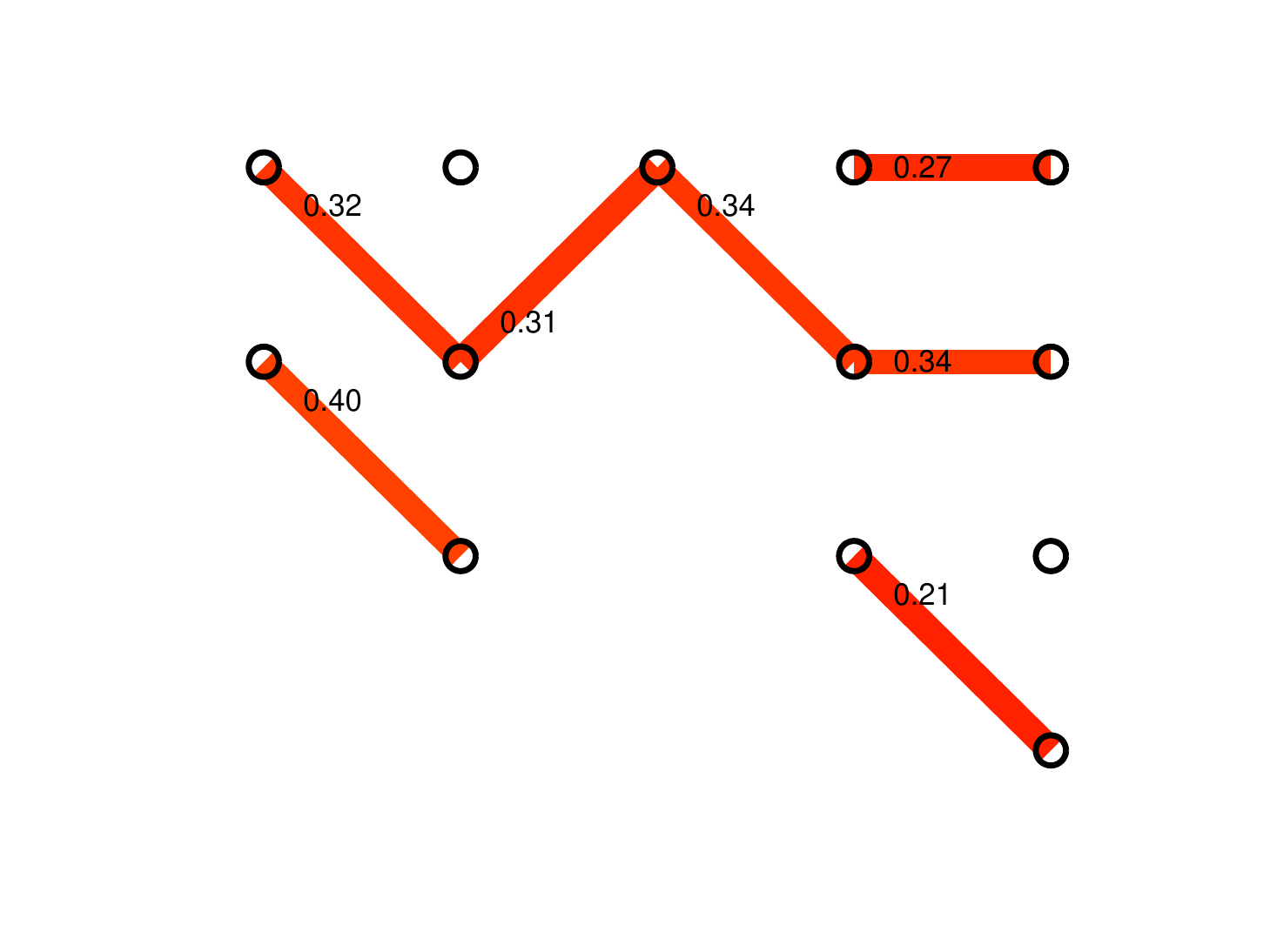}
	\end{center}
	\caption{The final lesion sequences found by processing the lesion graph in \Fig{matching-complete} using Algo.\ 1 in the paper. They are the same with the ground-truth in \Fig{matching-seq}.}
	\label{fig:matching-decomp}
\end{figure}

To provide a intuitive illustration of the lesion matching task, we show lesions of a sample patient in \Fig{matching-seq}, with their lesion graph in \Fig{matching-complete} and the final extracted lesion sequences in \Fig{matching-decomp}. We show that the lesion graph and Algo.\ 1 in the paper can be used to accurately match lesions in multiple studies.

\section*{Acknowledgments}
This research was supported by the Intramural Research Program of the NIH Clinical Center. We thank NVIDIA for the donation of GPU cards.

{\small
\bibliographystyle{ieee}
\bibliography{final}

\begin{thebibliography}{10}\itemsep=-1pt

\bibitem{belkin2006manifold}
M.~Belkin, P.~Niyogi, and V.~Sindhwani.
\newblock {Manifold regularization: A geometric framework for learning from
  labeled and unlabeled examples}.
\newblock {\em Journal of machine learning research}, 7(Nov):2399--2434, 2006.

\bibitem{bellet2013survey}
A.~Bellet, A.~Habrard, and M.~Sebban.
\newblock A survey on metric learning for feature vectors and structured data.
\newblock {\em arXiv preprint arXiv:1306.6709}, 2013.

\bibitem{bromley1994signature}
J.~Bromley, I.~Guyon, Y.~LeCun, E.~S{\"a}ckinger, and R.~Shah.
\newblock Signature verification using a ``siamese'' time delay neural network.
\newblock In {\em Advances in Neural Information Processing Systems}, pages
  737--744, 1994.

\bibitem{Cai2018MICCAI}
J.~Cai, Y.~Tang, L.~Lu, A.~P. Harrison, K.~Yan, J.~Xiao, L.~Yang, and R.~M.
  Summers.
\newblock {Accurate Weakly-Supervised Deep Lesion Segmentation using
  Large-Scale Clinical Annotations: Slice-Propagated 3D Mask Generation from 2D
  RECIST}.
\newblock In {\em MICCAI}, 2018.

\bibitem{Chen2017beyond}
W.~Chen, X.~Chen, J.~Zhang, and K.~Huang.
\newblock {Beyond triplet loss: a deep quadruplet network for person
  re-identification}.
\newblock In {\em CVPR}, 2017.

\bibitem{chen2015webly}
X.~Chen and A.~Gupta.
\newblock Webly supervised learning of convolutional networks.
\newblock In {\em Proceedings of the IEEE International Conference on Computer
  Vision}, pages 1431--1439, 2015.

\bibitem{Cheng2016CAD}
J.-Z. Cheng, D.~Ni, Y.-H. Chou, J.~Qin, C.-M. Tiu, Y.-C. Chang, C.-S. Huang,
  D.~Shen, and C.-M. Chen.
\newblock {Computer-Aided Diagnosis with Deep Learning Architecture:
  Applications to Breast Lesions in US Images and Pulmonary Nodules in CT
  Scans}.
\newblock {\em Scientific Reports}, 6(1):24454, jul 2016.

\bibitem{Clark2013tcia}
K.~Clark, B.~Vendt, K.~Smith, J.~Freymann, J.~Kirby, P.~Koppel, S.~Moore,
  S.~Phillips, D.~Maffitt, M.~Pringle, L.~Tarbox, and F.~Prior.
\newblock {The Cancer Imaging Archive (TCIA): Maintaining and Operating a
  Public Information Repository}.
\newblock {\em Journal of Digital Imaging}, 26(6):1045--1057, dec 2013.

\bibitem{cornegruta2016modeling}
S.~Cornegruta, R.~Bakewell, S.~Withey, and G.~Montana.
\newblock Modelling radiological language with bidirectional long short-term
  memory networks.
\newblock {\em arXiv preprint arXiv:1609.08409}, 2016.

\bibitem{Feifei09ImageNet}
J.~Deng, W.~Dong, R.~Socher, L.~J. Li, K.~Li, and L.~Fei-Fei.
\newblock {ImageNet: A large-scale hierarchical image database}.
\newblock In {\em 2009 IEEE Conference on Computer Vision and Pattern
  Recognition}, pages 248--255, jun 2009.

\bibitem{Eisen09RECIST}
E.~Eisenhauer, P.~Therasse, J.~Bogaerts, L.~H. Schwartz, D.~Sargent, R.~Ford,
  J.~Dancey, S.~Arbuck, S.~Gwyther, M.~Mooney, and Others.
\newblock {New response evaluation criteria in solid tumours: revised RECIST
  guideline (version 1.1)}.
\newblock {\em European journal of cancer}, 45(2):228--247, 2009.

\bibitem{esteva2017dermatologist}
A.~Esteva, B.~Kuprel, R.~A. Novoa, J.~Ko, S.~M. Swetter, H.~M. Blau, and
  S.~Thrun.
\newblock Dermatologist-level classification of skin cancer with deep neural
  networks.
\newblock {\em Nature}, 542(7639):115--118, 2017.

\bibitem{everingham2010pascal}
M.~Everingham, L.~Van~Gool, C.~K. Williams, J.~Winn, and A.~Zisserman.
\newblock The pascal visual object classes (voc) challenge.
\newblock {\em International journal of computer vision}, 88(2):303--338, 2010.

\bibitem{gidaris2015object}
S.~Gidaris and N.~Komodakis.
\newblock Object detection via a multi-region and semantic segmentation-aware
  cnn model.
\newblock In {\em Proceedings of the IEEE International Conference on Computer
  Vision}, pages 1134--1142, 2015.

\bibitem{Girshick2015fast}
R.~Girshick.
\newblock {Fast r-cnn}.
\newblock In {\em Proceedings of the IEEE international conference on computer
  vision}, pages 1440--1448, 2015.

\bibitem{Greenspan2016guest}
H.~Greenspan, B.~van Ginneken, and R.~M. Summers.
\newblock {Guest Editorial Deep Learning in Medical Imaging: Overview and
  Future Promise of an Exciting New Technique}.
\newblock {\em IEEE Transactions on Medical Imaging}, 35(5):1153--1159, may
  2016.

\bibitem{hofmanninger2016unsupervised}
J.~Hofmanninger, M.~Krenn, M.~Holzer, T.~Schlegl, H.~Prosch, and G.~Langs.
\newblock {Unsupervised identification of clinically relevant clusters in
  routine imaging data}.
\newblock In {\em International Conference on Medical Image Computing and
  Computer-Assisted Intervention}, pages 192--200. Springer, 2016.

\bibitem{Hong2008auto}
H.~Hong, J.~Lee, and Y.~Yim.
\newblock {Automatic lung nodule matching on sequential CT images}.
\newblock {\em Computers in Biology and Medicine}, 38(5):623--634, may 2008.

\bibitem{hu2017tiny}
P.~Hu and D.~Ramanan.
\newblock {Finding Tiny Faces}.
\newblock In {\em CVPR}, 2017.

\bibitem{krause2016unreasonable}
J.~Krause, B.~Sapp, A.~Howard, H.~Zhou, A.~Toshev, T.~Duerig, J.~Philbin, and
  L.~Fei-Fei.
\newblock The unreasonable effectiveness of noisy data for fine-grained
  recognition.
\newblock In {\em European Conference on Computer Vision}, pages 301--320.
  Springer, 2016.

\bibitem{lee2013pseudo}
D.-H. Lee.
\newblock Pseudo-label: The simple and efficient semi-supervised learning
  method for deep neural networks.
\newblock In {\em Workshop on Challenges in Representation Learning, ICML},
  volume~3, page~2, 2013.

\bibitem{Li2018review}
Z.~Li, X.~Zhang, H.~M{\"{u}}ller, and S.~Zhang.
\newblock {Large-scale retrieval for medical image analytics: A comprehensive
  review}.
\newblock {\em Medical Image Analysis}, 43:66--84, jan 2018.

\bibitem{lin2014microsoft}
T.-Y. Lin, M.~Maire, S.~Belongie, J.~Hays, P.~Perona, D.~Ramanan,
  P.~Doll{\'{a}}r, and C.~L. Zitnick.
\newblock {Microsoft coco: Common objects in context}.
\newblock In {\em European conference on computer vision}, pages 740--755.
  Springer, 2014.

\bibitem{Litjens2017survey}
G.~Litjens, T.~Kooi, B.~E. Bejnordi, A.~A.~A. Setio, F.~Ciompi, M.~Ghafoorian,
  J.~A. van~der Laak, B.~van Ginneken, and C.~I. S{\'{a}}nchez.
\newblock {A survey on deep learning in medical image analysis}.
\newblock {\em Medical Image Analysis}, 42:60--88, dec 2017.

\bibitem{Manning2008intro}
C.~D. Manning, P.~Raghavan, and H.~Schütze.
\newblock {\em {Introduction to information retrieval}}.
\newblock Cambridge University Press, 2008.

\bibitem{moltz2012workflow}
J.~H. Moltz, M.~D'Anastasi, A.~Kie{\ss}ling, D.~P. {Dos Santos},
  C.~Sch{\"{u}}lke, and H.-O. Peitgen.
\newblock {Workflow-centred evaluation of an automatic lesion tracking software
  for chemotherapy monitoring by CT}.
\newblock {\em European radiology}, 22(12):2759--2767, 2012.

\bibitem{moltz2009general}
J.~H. Moltz, M.~Schwier, and H.-O. Peitgen.
\newblock A general framework for automatic detection of matching lesions in
  follow-up ct.
\newblock In {\em Biomedical Imaging: From Nano to Macro, 2009. ISBI'09. IEEE
  International Symposium on}, pages 843--846. IEEE, 2009.

\bibitem{oh2016deep}
H.~Oh~Song, Y.~Xiang, S.~Jegelka, and S.~Savarese.
\newblock Deep metric learning via lifted structured feature embedding.
\newblock In {\em Proceedings of the IEEE Conference on Computer Vision and
  Pattern Recognition}, pages 4004--4012, 2016.

\bibitem{Ramos2016content}
J.~Ramos, T.~T. J.~P. Kockelkorn, I.~Ramos, R.~Ramos, J.~Grutters, M.~A.
  Viergever, B.~van Ginneken, and A.~Campilho.
\newblock {Content-Based Image Retrieval by Metric Learning From Radiology
  Reports: Application to Interstitial Lung Diseases}.
\newblock {\em IEEE Journal of Biomedical and Health Informatics},
  20(1):281--292, jan 2016.

\bibitem{schroff2015facenet}
F.~Schroff, D.~Kalenichenko, and J.~Philbin.
\newblock Facenet: A unified embedding for face recognition and clustering.
\newblock In {\em Proceedings of the IEEE Conference on Computer Vision and
  Pattern Recognition}, pages 815--823, 2015.

\bibitem{setio2017validation}
A.~A.~A. Setio, A.~Traverso, T.~De~Bel, M.~S. Berens, C.~van~den Bogaard,
  P.~Cerello, H.~Chen, Q.~Dou, M.~E. Fantacci, B.~Geurts, et~al.
\newblock Validation, comparison, and combination of algorithms for automatic
  detection of pulmonary nodules in computed tomography images: the luna16
  challenge.
\newblock {\em Medical Image Analysis}, 42:1--13, 2017.

\bibitem{sevenster2015improved}
M.~Sevenster, A.~R. Travis, R.~K. Ganesh, P.~Liu, U.~Kose, J.~Peters, and P.~J.
  Chang.
\newblock Improved efficiency in clinical workflow of reporting measured
  oncology lesions via pacs-integrated lesion tracking tool.
\newblock {\em American Journal of Roentgenology}, 204(3):576--583, 2015.

\bibitem{shin2016interleaved}
H.-C. Shin, L.~Lu, L.~Kim, A.~Seff, J.~Yao, and R.~Summers.
\newblock {Interleaved text/image deep mining on a large-scale radiology
  database for automated image interpretation}.
\newblock {\em Journal of Machine Learning Research}, 17(1-31):2, 2016.

\bibitem{Shin2016tmi}
H.-C. Shin, H.~R. Roth, M.~Gao, L.~Lu, Z.~Xu, I.~Nogues, J.~Yao, D.~Mollura,
  and R.~M. Summers.
\newblock {Deep Convolutional Neural Networks for Computer-Aided Detection: CNN
  Architectures, Dataset Characteristics and Transfer Learning}.
\newblock {\em IEEE Transactions on Medical Imaging}, 35(5):1285--1298, may
  2016.

\bibitem{silva2011fast}
J.~S. Silva, J.~Cancela, and L.~Teixeira.
\newblock Fast volumetric registration method for tumor follow-up in pulmonary
  ct exams.
\newblock {\em Journal of Applied Clinical Medical Physics}, 12(2):362--375,
  2011.

\bibitem{Simonyan2015Vgg}
K.~Simonyan and A.~Zisserman.
\newblock {Very deep convolutional networks for large-scale image recognition}.
\newblock In {\em ICLR 2015}, 2015.

\bibitem{sohn16npair}
K.~Sohn.
\newblock {Improved Deep Metric Learning with Multi-class N-pair Loss
  Objective}.
\newblock In {\em Neural Information Processing Systems}, pages 1--9, 2016.

\bibitem{Son2017multi}
J.~Son, M.~Baek, M.~Cho, and B.~Han.
\newblock {Multi-Object Tracking with Quadruplet Convolutional Neural
  Networks}.
\newblock In {\em Proceedings of the IEEE Conference on Computer Vision and
  Pattern Recognition}, pages 5620--5629, 2017.

\bibitem{song17deep}
H.~O. Song, S.~Jegelka, V.~Rathod, and K.~Murphy.
\newblock Deep metric learning via facility location.
\newblock In {\em IEEE CVPR}, 2017.

\bibitem{Tajbakhsh2016tmi}
N.~Tajbakhsh, J.~Y. Shin, S.~R. Gurudu, R.~T. Hurst, C.~B. Kendall, M.~B.
  Gotway, and J.~Liang.
\newblock {Convolutional Neural Networks for Medical Image Analysis: Full
  Training or Fine Tuning?}
\newblock {\em IEEE Transactions on Medical Imaging}, 35(5):1299--1312, may
  2016.

\bibitem{teramoto2016automated}
A.~Teramoto, H.~Fujita, O.~Yamamuro, and T.~Tamaki.
\newblock Automated detection of pulmonary nodules in pet/ct images: Ensemble
  false-positive reduction using a convolutional neural network technique.
\newblock {\em Medical physics}, 43(6):2821--2827, 2016.

\bibitem{tsochatzidis2017computer}
L.~Tsochatzidis, K.~Zagoris, N.~Arikidis, A.~Karahaliou, L.~Costaridou, and
  I.~Pratikakis.
\newblock {Computer-aided diagnosis of mammographic masses based on a
  supervised content-based image retrieval approach}.
\newblock {\em Pattern Recognition}, 2017.

\bibitem{Maaten2014tsne}
L.~van~der Maaten.
\newblock {Accelerating t-SNE using Tree-Based Algorithms}.
\newblock {\em Journal of Machine Learning Research}, 15:3221--3245, 2014.

\bibitem{vivanti2015automatic}
R.~Vivanti.
\newblock Automatic liver tumor segmentation in follow-up ct studies using
  convolutional neural networks.
\newblock In {\em Proc. Patch-Based Methods in Medical Image Processing
  Workshop}, 2015.

\bibitem{wang2017unsupervised}
X.~Wang, L.~Lu, H.-C. Shin, L.~Kim, M.~Bagheri, I.~Nogues, J.~Yao, and R.~M.
  Summers.
\newblock Unsupervised joint mining of deep features and image labels for
  large-scale radiology image categorization and scene recognition.
\newblock In {\em Applications of Computer Vision (WACV), 2017 IEEE Winter
  Conference on}, pages 998--1007. IEEE, 2017.

\bibitem{Wang2017ChestXray8}
X.~Wang, Y.~Peng, L.~Lu, Z.~Lu, M.~Bagheri, and R.~M. Summers.
\newblock {ChestX-ray8: Hospital-scale Chest X-ray Database and Benchmarks on
  Weakly-Supervised Classification and Localization of Common Thorax Diseases}.
\newblock In {\em CVPR}, may 2017.

\bibitem{Wang2017zoomin}
Z.~Wang, Y.~Yin, J.~Shi, W.~Fang, H.~Li, and X.~Wang.
\newblock {\em {Zoom-in-Net: Deep Mining Lesions for Diabetic Retinopathy
  Detection}}, pages 267--275.
\newblock Springer International Publishing, 2017.

\bibitem{Wei2016sim}
G.~Wei, H.~Ma, W.~Qian, and M.~Qiu.
\newblock {Similarity measurement of lung masses for medical image retrieval
  using kernel based semisupervised distance metric}.
\newblock {\em Medical Physics}, 43(12):6259--6269, nov 2016.

\bibitem{zhang2016online}
H.~Zhang, X.~Shang, W.~Yang, H.~Xu, H.~Luan, and T.-S. Chua.
\newblock Online collaborative learning for open-vocabulary visual classifiers.
\newblock In {\em Proceedings of the IEEE Conference on Computer Vision and
  Pattern Recognition}, pages 2809--2817, 2016.

\bibitem{Zhang2016embedding}
X.~Zhang, F.~Zhou, Y.~Lin, and S.~Zhang.
\newblock Embedding label structures for fine-grained feature representation.
\newblock In {\em Proceedings of the IEEE Conference on Computer Vision and
  Pattern Recognition}, pages 1114--1123, 2016.

\bibitem{zhang2017mdnet}
Z.~Zhang, Y.~Xie, F.~Xing, M.~McGough, and L.~Yang.
\newblock {MDNet: A Semantically and Visually Interpretable Medical Image
  Diagnosis Network}.
\newblock In {\em CVPR}, 2017.

\bibitem{zhu2002learning}
X.~Zhu and Z.~Ghahramani.
\newblock Learning from labeled and unlabeled data with label propagation.
\newblock {\em Technical Report CMU-CALD-02-107, Carnegie Mellon University},
  2002.

\end{thebibliography}
}

\end{document}